\documentclass[english]{pfia}

\usepackage{caption}
\usepackage{subcaption}
\usepackage{graphicx}
\usepackage{amsmath}
\usepackage{amssymb}
\usepackage{hyperref}
\hypersetup{colorlinks=true}
\usepackage{float}
\usepackage{bm}

\usepackage[page]{appendix}

\newcommand{\vect}[1]{\pmb{#1}}

\newif\ifdetails
\newcommand{\inDetailledVersion}[1]{\ifdetails #1\fi}

\newif\ifappendix
\newcommand{\includeAppendix}[1]{\ifappendix #1\fi}

\title{\textbf{The Epistemic Uncertainty Hole:\\ an issue of Bayesian Neural Networks}}

\author{M. Fellaji\textsuperscript{1,2}, F. Pennerath\textsuperscript{1,2}\\[6pt]
\textsuperscript{1} Laboratoire lorrain de recherche en informatique et ses applications, LORIA\\
\textsuperscript{2} CentraleSupélec, CS}

\date{
\href{mailto:mohammed.fellaji@centralesupelec.fr}{mohammed.fellaji@centralesupelec.fr}\; 
\; \; \; \href{mailto:frederic.pennerath@loria.fr}{frederic.pennerath@loria.fr}
}

\begin{document}

\maketitle

% ------------------------------------------
% RÉSUMÉS ET MOTS-CLÉS
% ------------------------------------------

\begin{abstract}
  \emph{Bayesian Deep Learning} (BDL) gives access not only to \emph{aleatoric uncertainty}, as standard neural networks already do, but also to \emph{epistemic uncertainty}, a measure of confidence a model has in its own predictions.
  In this article we show through experiments that the evolution of epistemic uncertainty metrics regarding the model size and the size of the training set, goes against theoretical expectations.
  More precisely, we observe that the epistemic uncertainty collapses literally in the presence of large models and sometimes also of little training data, while we expect the exact opposite behaviour.
  This phenomenon, which we call "\emph{epistemic uncertainty hole}", is all the more problematic as it undermines the entire applicative potential of BDL, which is based precisely on the use of epistemic uncertainty.
  As an example, we evaluate the practical consequences of this uncertainty hole on one of the main applications of BDL, namely the \emph{detection of out-of-distribution samples}.
%We study in depth the effect of the size of the training set on the resulting uncertainties of the models. We will focus mainly on the epistemic uncertainty, which should decrease in the presence of more datapoints. However, contrarily to its definition, we noticed that it does not decrease always when increasing the size of the training set. To have a better understanding of this observation, we will also include the effect of the size of the model in our analysis. Furthermore, we will make the distinction between the good classified and the misclassified examples. Finally, we will test the trained models on Out-of-Distribution (OOD) samples.
\end{abstract}


\begin{resume}
  L'\emph{apprentissage profond bayésien} (BDL) donne accès non seulement à l'\emph{incertitude aléatoire}, comme le font déjà les réseaux neuronaux standards, mais aussi à l'\emph{incertitude épistémique}, une mesure de la confiance qu'a un modèle dans ses propres prédictions.
  Dans cet article, nous montrons par l'expérience que l'évolution des mesures d'incertitude épistémique en fonction de la taille du modèle et de la taille du jeu d'apprentissage va à l'encontre de ce que la théorie prévoit.
  Nous observons ainsi que l'incertitude épistémique s'effondre littéralement en présence de grands modèles et parfois aussi de peu de données d'entraînement, alors que nous nous attendons au comportement inverse.
  Ce phénomène, que nous nommons "\emph{trou d'incertitude épistémique}", est d'autant plus problématique qu'il sape  le potentiel applicatif du BDL, puisque ce dernier repose précisément sur l'exploitation de l'incertitude épistémique.
  A titre d'exemple nous évaluons les conséquences pratiques de ce trou d'incertitude sur l'une des principales applications du BDL, à savoir la \emph{détection d'échantillons hors-distribution}.
\end{resume}

\begin{motscles}
Réseaux de neurones bayésiens, incertitude épistémique, calibration, détection hors-distribution, ensembles de modèles.
\end{motscles}

% ------------------------------------------
% CORPS DE L'ARTICLE
% ------------------------------------------

\section{Introduction}

In many applications of Machine Learning, optimizing solely the performance metrics of the predictive model, such as the accuracy, can result in overconfident interpretations of erroneous outcomes, and thus, hazardous decisions in case of critical domains. Therefore, being able to map the model outputs to some uncertainty quantification metrics, if well calibrated, is essential from a decision making point of view. 
%In addition, it could be hard to collect a large set of samples to train the deep learning models. .. \cite{mackay1992bayesian,} 
When dealing with Deep Learning models, \emph{Bayesian Deep Learning} (BDL)~\cite{mackay_practical_1992, neal_bayesian_1996, wilson_bayesian_2020, lakshminarayanan_simple_2017, blundell_weight_2015}, i.e. the application of \emph{Bayesian inference} to deep neural networks, appears to be one of the keys to estimate such well-calibrated uncertainties.
%is a rapidly growing field, judging by its adoption in an increasing number of applications.

In statistics, Bayesian inference is known to have unique assets, which classical point estimators (MLE, MAP, etc) do not have, one of which is its unique ability to measure \emph{epistemic uncertainty} (sometimes also called model uncertainty)~\cite{Gal2016Uncertainty, kwon_uncertainty_2020}. This form of uncertainty should not be confused with the \emph{aleatoric uncertainty} (or data uncertainty) intrinsic to the problem treated (e.g. superposition of the classes), which is estimated by modelling the distribution $P(Y \,|\,X, \theta)$ of the output $Y$ of the model conditionally on the inputs $X$ and the parameters $\theta$ of the model. In comparison, the epistemic uncertainty reflects the lack of knowledge on the value of the model's parameters due to the observation of a limited number of training examples, an uncertainty conveyed by the \emph{posterior distribution} $P(\theta\,|\, \mathcal{D})$ of the parameters, resulting from the conditioning of a \emph{prior distribution} by the training dataset $\mathcal{D}$. This posterior distribution on parameters in turn induces a \emph{posterior predictive distribution} on the output $P(Y \,|\, X) = \int P(Y \,|\, X, \theta) \, P(\theta \,|\, \mathcal{D}) \, d\theta$, which is necessarily more uncertain (i.e. of higher entropy) than if the parameters took a single known value, as is the case with point estimators. 
This quantifiable increase in uncertainty makes it possible to define and to compute the epistemic uncertainty for any model output.

%% Epistemic uncertainty vs OOD
Epistemic uncertainty can then be exploited in many applications. An important one is the identification of \emph{out-of-distribution examples} (OOD) underlying many real-world applications, like the detection of adversarial attacks~\cite{szegedy2013intriguing, smith2018understanding} and the design of safe and robust AI for critical applications~\cite{arnez2020comparison}. In theory, the level of aleatoric uncertainty of a prediction, as estimated by standard neural networks, does not tell anything about the possible OOD nature of the example. Only abnormally high levels of epistemic uncertainty, as predicted exclusively by Bayesian neural networks, can do so.
We could also quote other applications of epistemic uncertainty, in particular in the field of active learning~\cite{pmlr-v70-gal17a} and reinforcement learning~\cite{clements2019estimating}.

While from a theoretical point of view, Bayesian inference in general and Bayesian Deep Learning in particular are attractive, we show in this paper that in practice, the evolution of experimental measures of epistemic uncertainty with respect to the size of Bayesian deep neural networks (i.e. the number of parameters) and the size of the training set, goes against expectations and what theory predicts. 
More precisely, we observe that the epistemic uncertainty collapses literally in the presence of a strongly parameterized model and sometimes also of little training data, while we expect the exact opposite behaviour.
% MNIST~\cite{deng2012mnist} and CIFAR10~\cite{cifar10dataset}
We observe this phenomenon on several well-known prototypical experiments, including dense networks applied to the image datasets MNIST and CIFAR10, Convolutional Neural Networks (CNN) ResNet ~\cite{he_deep_2015} applied to CIFAR10 and two distinct fundamental types of BDL models (ensembles~\cite{lakshminarayanan_simple_2017} and MC-Dropout~\cite{gal_dropout_2016}).
This phenomenon, which we call the "\emph{epistemic uncertainty hole}", is all the more problematic as it undermines the entire applicative potential of the BDL, which is based precisely on the exploitation of epistemic uncertainty.

The first objective and main contribution of this paper is to bring to light this "epistemic uncertainty hole" in a rigorous way and on several iconic experiments.
The second objective is to measure what this epistemic uncertainty hole implies on an essential BDL application, namely the detection of OOD examples.
%Finally, the last objective, more exploratory, is to propose first insights about the reason of this phenomenon.

The plan of the article is structured as follows.
In section~\ref{sec:BDL}, the basics of BDL are recalled, in particular the way the epistemic uncertainty is computed in the case of a supervised classification problem.
In section~\ref{sec:experiments}, the results of different experiments are presented, highlighting the "epistemic uncertainty hole".
In section~\ref{sec:OOD}, the negative consequences of the "hole" on the detection of OOD examples are evaluated.
Finally, the section~\ref{sec:conclusion} concludes on the research perspectives raised by the problem thus depicted.

\begin{figure*}[htbp]
     \centering
     % \includegraphics[width=\textwidth]{figures/ensemble/resnet18_noWD/mutual_information.pdf}
     % \caption{Box-plots of the epistemic uncertainty for the evaluations of the ensemble of ResNet18 models trained on CIFAR10 and tested on the test set of CIFAR10 (\textbf{ID}: In-Distribution examples) and on the test set of SVHN (\textbf{OOD}: Out-Of-Distribution examples). On the x-axis we have the number of examples used to train the models and the normalized epistemic uncertainty on the y-axis. \textbf{ID-mis} represents the misclassified examples from the ID set, \textbf{ID-all} is for the entire ID set and \textbf{ID-good} are the ID examples correctly classified by the ensemble.} 
     \includegraphics[width=\textwidth]{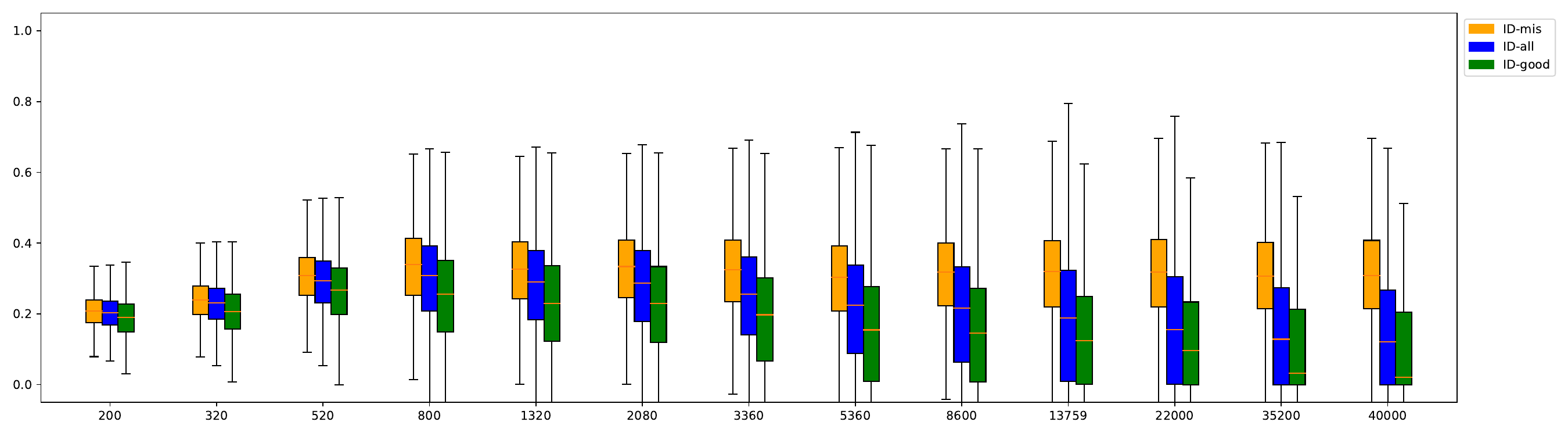}
     \caption{Box-plots of the epistemic uncertainty for the evaluations of the ensemble of ResNet18 models trained on CIFAR10 and tested on the test set of CIFAR10 (\textbf{ID}: In-Distribution examples). On the x-axis we have the number of examples used to train the models and the normalized epistemic uncertainty on the y-axis. \textbf{ID-mis} represents the misclassified examples from the ID set, \textbf{ID-all} is for the entire ID set and \textbf{ID-good} are the ID examples correctly classified by the ensemble.} 
     \label{fig:resnet18-cifar10}
\end{figure*}

\section{On Epistemic Uncertainty~\label{sec:BDL}}

In the following, we focus on neural networks addressing supervised classification problems.
In such problems, neural networks produce as an output a categorical distribution $P(Y \, | \, \vect{x}, \vect{w})$ of the class variable $Y \in \{ 1,\dots, C \}$ predicted from input features\footnote{Thereafter bold symbols refer to vectors or more general tensors.} $\vect{x}$.
Vector $\vect{w}$ refers to the set of optimizable parameters (i.e weights and biases).
In standard network, the MAP estimator $\hat{\vect{w}}_{MAP}$ is computed by minimizing a regularized cross-entropy loss.
However, in Bayesian networks, parameters are processed as a random variable $\vect{W}$ following some prior distribution $p(\vect{W})$.
Training consists in inferring the posterior distribution $p(\vect{W} \, | \, \mathcal{D})$ conditioned on the training dataset $\mathcal{D} = \{(\vect{x}_i, y_i) \}$ thanks to Bayes' rule
\[
p(\vect{W}\,|\,\mathcal{D}) = \frac{p(\mathcal{D}|\vect{W}) \, p(\vect{W})}{p(\mathcal{D})} \quad .
   \label{eq:bayes}
\]
This inference cannot be exact for complex models like neural networks; hence approximate techniques are needed.
\emph{Variational inference} along with stochastic gradient descent based on the reparameterization trick~\cite{kingma2013auto} are used to infer a simpler approximated posterior $q_{\vect{\theta}}(\vect{W}) \approx p(\vect{W}\,|\,\mathcal{D})$ where $\vect{\theta}$ is the set of variational parameters for some predefined family of distributions over the network weights $\vect{W}$.
Many different families of variational distributions $q_{\vect{\theta}}$ have been proposed in the field of BDL. We considered here two prominent options, namely MC dropout~\cite{gal_dropout_2016} and model ensembles~\cite{lakshminarayanan_simple_2017}.
% kingma paper for the reparameterization trick

Whatever the family of proxies $q_{\vect{\theta}}$ retained, the variational parameters $\vect{\theta}$, once learnt, are used to estimate the posterior predictive distribution for some new input $\vect{x}$, using Monte Carlo, i.e by sampling $K$ parameter sets $\vect{w}_i$ from the approximated posterior:
\begin{eqnarray*}
   p(Y\,|\,\vect{x},\mathcal{D}) & \approx & \int p(Y\,|\,\vect{x},\vect{w}) \, q_{\vect{\theta}}(\vect{w}) \, d\vect{w} \\
   & \approx &  \frac{1}{K} \sum_{i=1}^K p(Y\,|\,\vect{x},\vect{w}_i) \text{ with } \vect{w}_i \sim q_{\vect{\theta}}(\vect{W})
\end{eqnarray*}
In addition to the output distribution, metrics of predictive, aleatoric, and epistemic uncertainties can be assessed~\cite{Gal2016Uncertainty}:
\begin{itemize}
\item \emph{Predictive uncertainty} is the total uncertainty of the model for some sample $\vect{x}$, quantified as the entropy of the posterior predictive distribution, i.e.
\begin{eqnarray*}
  \mathcal{U}_{total}(\vect{x}) & = & \mathcal{H}(Y\,|\,\vect{x},\mathcal{D}) \\
                                & = & -\sum_{y = 1}^C p(y\,|\,\vect{x},\mathcal{D}) \, \log\left(p(y\,|\,\vect{x},\mathcal{D})\right) \, .
\end{eqnarray*}
\item \emph{Aleatoric uncertainty} is the uncertainty intrinsic to the data due to hidden variables or measurement errors. It is defined as the average of the uncertainties as predicted by every sampled model, that is the conditional entropy of $Y$ given $\vect{W}$, i.e.
\begin{eqnarray*}
  \mathcal{U}_{aleat.}(\vect{x}) & = & \mathcal{H}(Y\,|\,\vect{x}, \vect{W},\mathcal{D})\\
                                 & = & \int \mathcal{H}(Y\,|\,\vect{x}, \vect{w}) \, q_{\vect{\theta}}(\vect{w}) \, d\vect{w} \\
                                 & \approx & \frac{1}{K} \sum_{i=1}^K \mathcal{H}(Y\,|\,\vect{x}, \vect{w}_i) \, .
\end{eqnarray*}
\item Finally \emph{epistemic uncertainty} is the difference between the predictive and aleatoric uncertainties, that is the mutual information between $Y$ and $\vect{W}$, i.e
\begin{eqnarray*}
  \mathcal{U}_{epist.}(\vect{x}) & = &  \mathcal{H}(Y\,|\,\vect{x},\mathcal{D}) - \mathcal{H}(Y\,|\,\vect{x}, \vect{W},\mathcal{D})\\
  & = & I(Y ; \vect{W} \,|\, \vect{x},\mathcal{D}) \, .
\end{eqnarray*}
\emph{Epistemic uncertainty} is thus the portion of uncertainty on $Y$ given $\vect{x}$ that is shared with $\vect{W}$ and therefore, that is reducible by observing more couples $(\vect{x}_i,y_i)$. Intuitively the epistemic uncertainty measures the surprise effect caused by a sample $\vect{x}$, i.e. the discrepancies between $\vect{x}$ and the training dataset.
\end{itemize}
All previous uncertainties are upper bounded by $\log(C)$ (i.e. the maximal entropy that a $C$-categorical distribution can reach). Consequently and for the sake of comparison, we divide uncertainty metrics by $\log(C)$ to consider normalized versions ranging from $0$ to $1$.

Given these considerations, several properties are expected from epistemic uncertainty:
\begin{itemize}
\item First, the average epistemic uncertainty $\bar{\mathcal{U}}_{epist.}$ computed on some test dataset
  is expected to decrease when the size of the training dataset $\mathcal{D}$ increases, as then, the test samples are more likely to be similar to the train dataset so that the  surprise effect should get lower on average.
\item Second, $\bar{\mathcal{U}}_{epist.}$ is expected to increase with the number of model parameters, i.e the dimension of $\vect{W}$, as the more parameters a model has, the more likely it is to fit the data in multiple ways.
Put another way, the posterior and thus the posterior predictive will tend to be flatter, making the epistemic uncertainty grow.
\item Finally the average epistemic uncertainty $\bar{\mathcal{U}}_{epist.}^{ood}$ computed on OOD samples (i.e samples that are distinctly different from the training examples) is expected to be significantly larger than uncertainty $\bar{\mathcal{U}}_{epist.}$ computed on \emph{in-distribution samples} (ID).
\end{itemize}
The next section explores experimentally the extent to which the two first requirements are met whereas section~\ref{sec:OOD} addresses the third expectation.

\section{The Hole of Epistemic Uncertainty~\label{sec:experiments}}
In this section, we illustrate from an experimental perspective the hole of epistemic uncertainty.
%Two benchmark datasets are used to train the models: \textbf{MNIST} and \textbf{CIFAR10}. 
\includeAppendix{For more details about the implementation, see the appendix~\ref{sec:imp}.}

%% epistemic uncertainties ID: (mnist vs cifar10) and (ensemble vs MC-Dropout)
\begin{figure*}[htbp]
     \centering
     \begin{subfigure}[b]{0.245\textwidth}
         \centering
         \includegraphics[width=\textwidth]{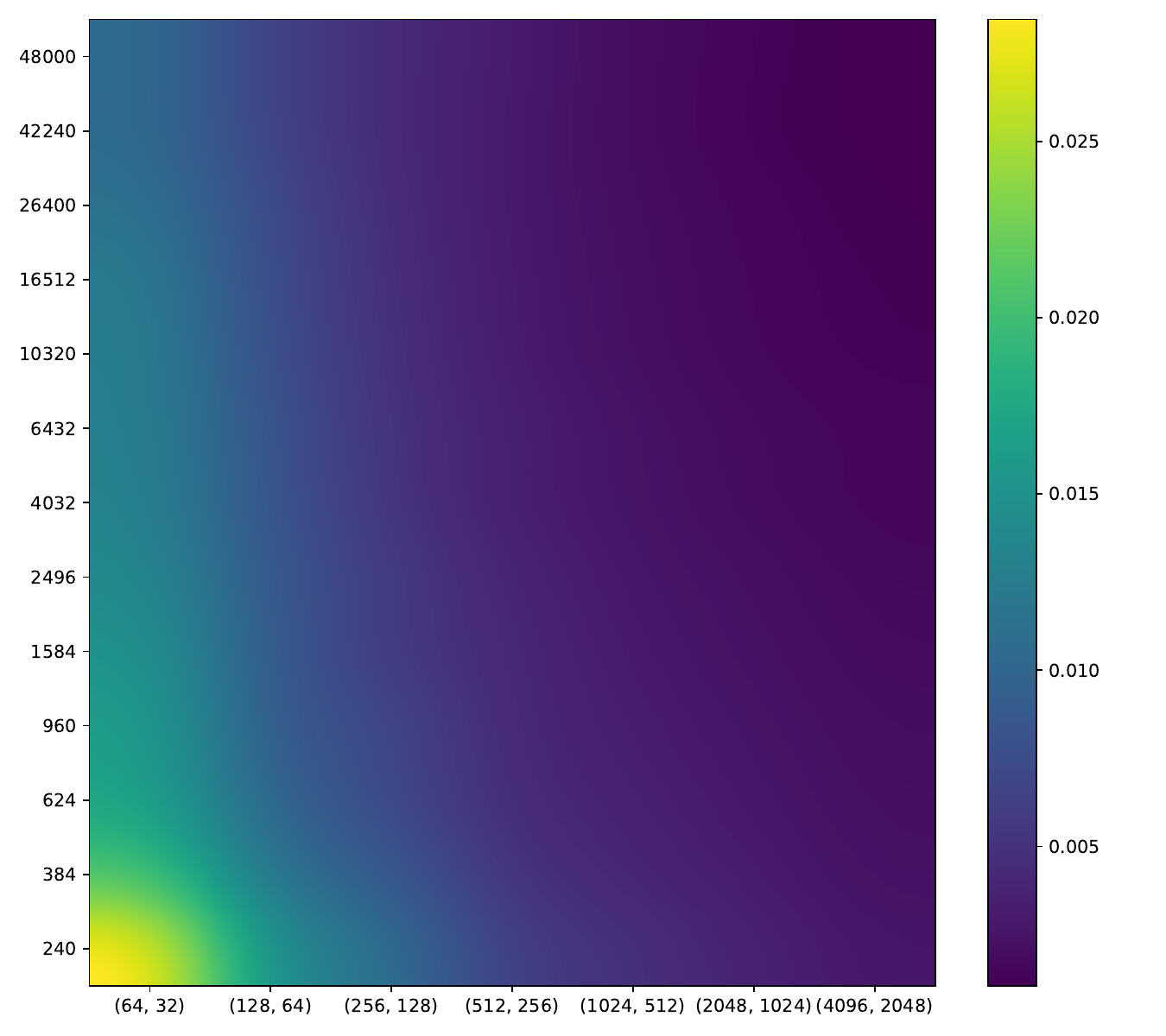}
         \caption{MNIST - Ensemble}
         \label{fig:epis-uncer-id-mnist-ensemble}
     \end{subfigure}
     \hfill
     \begin{subfigure}[b]{0.245\textwidth}
         \centering
         \includegraphics[width=\textwidth]{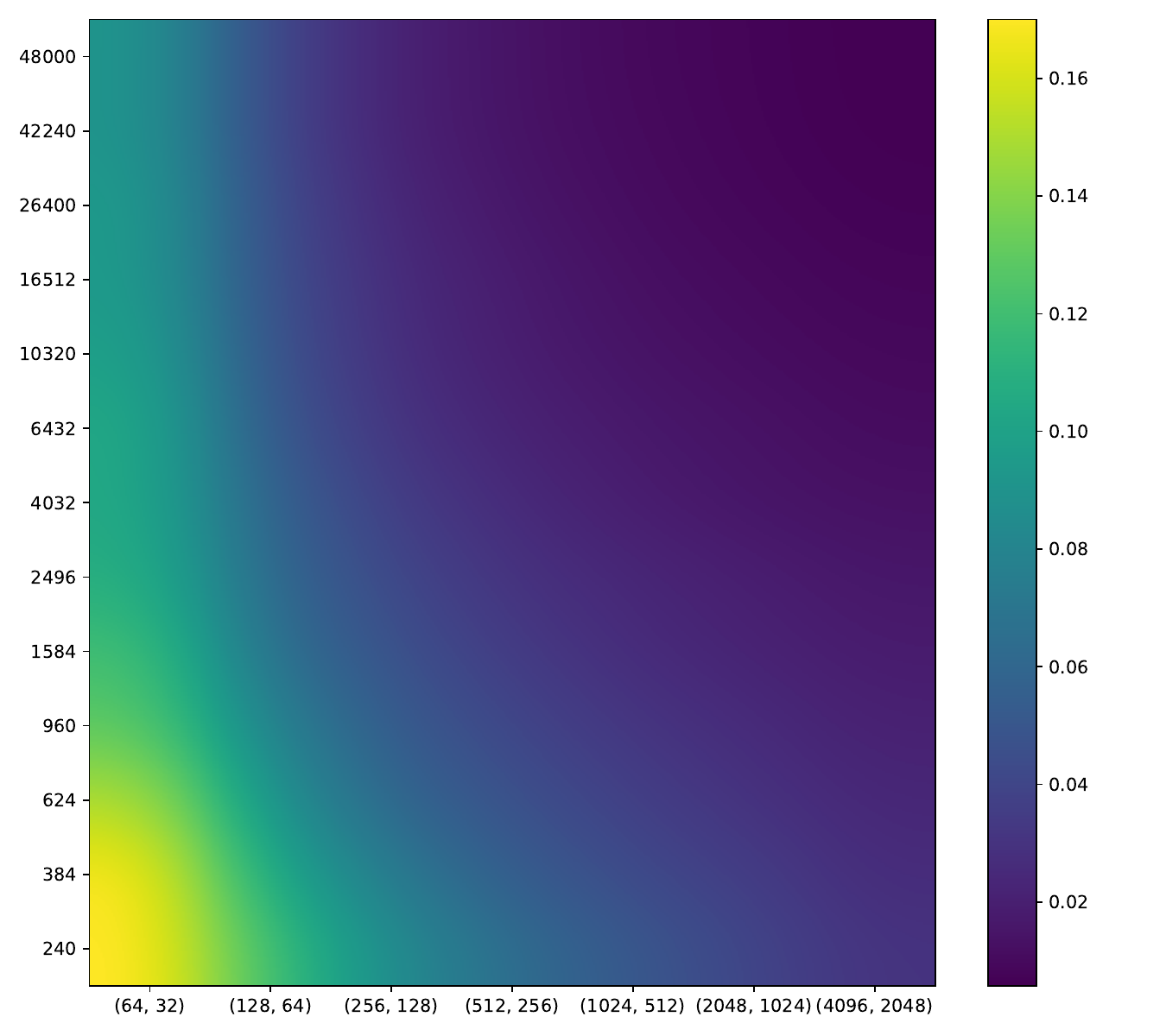}
         \caption{MNIST - MC-Dropout}
         \label{fig:epis-uncer-id-mnist-mc-dropout}
     \end{subfigure}
     \hfill
     \begin{subfigure}[b]{0.245\textwidth}
         \centering
         \includegraphics[width=\textwidth]{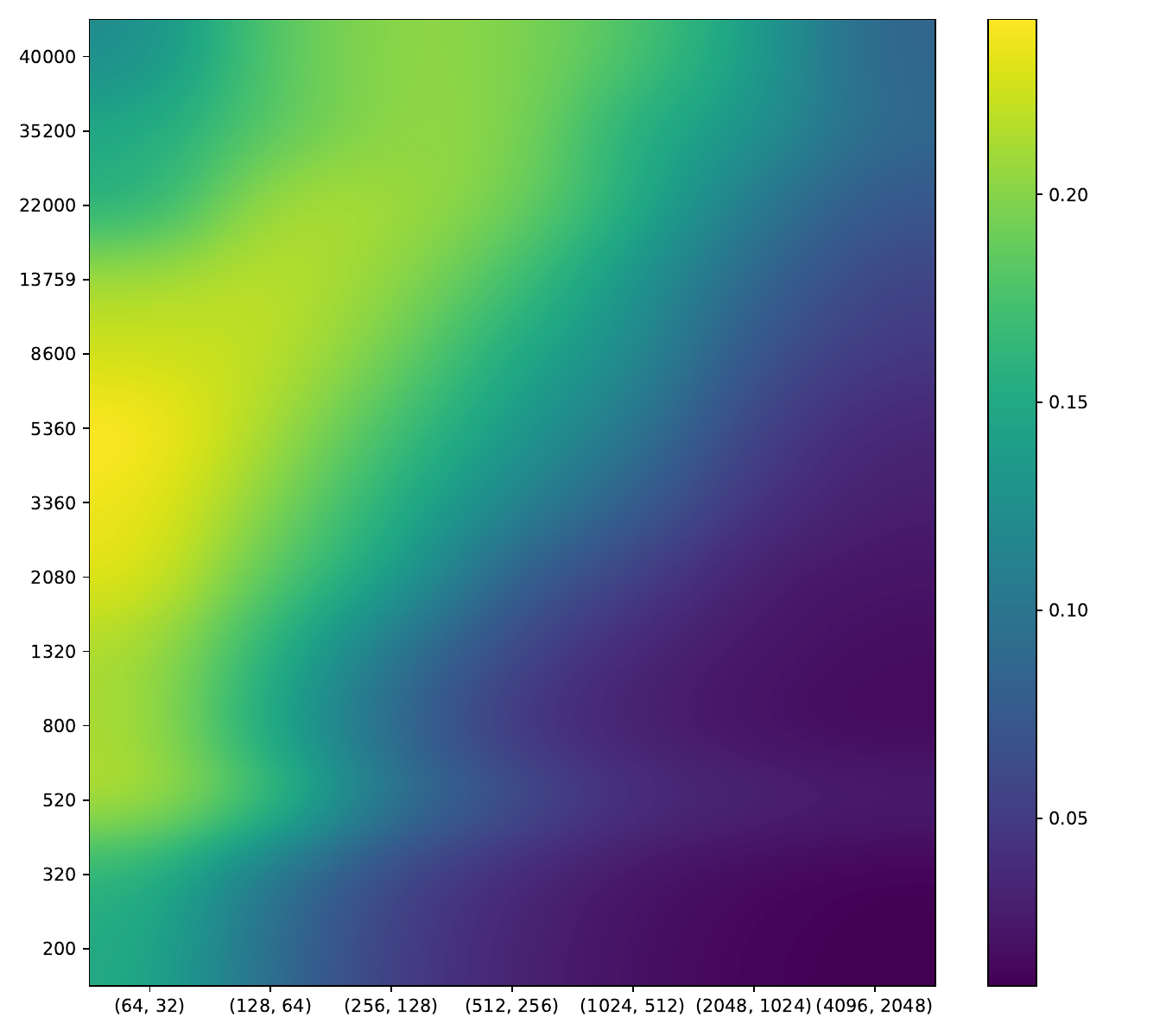}
         \caption{CIFAR10 - Ensemble}
         \label{fig:epis-uncer-id-cifar10-ensemble}
     \end{subfigure}
     \hfill
     \begin{subfigure}[b]{0.245\textwidth}
         \centering
         \includegraphics[width=\textwidth]{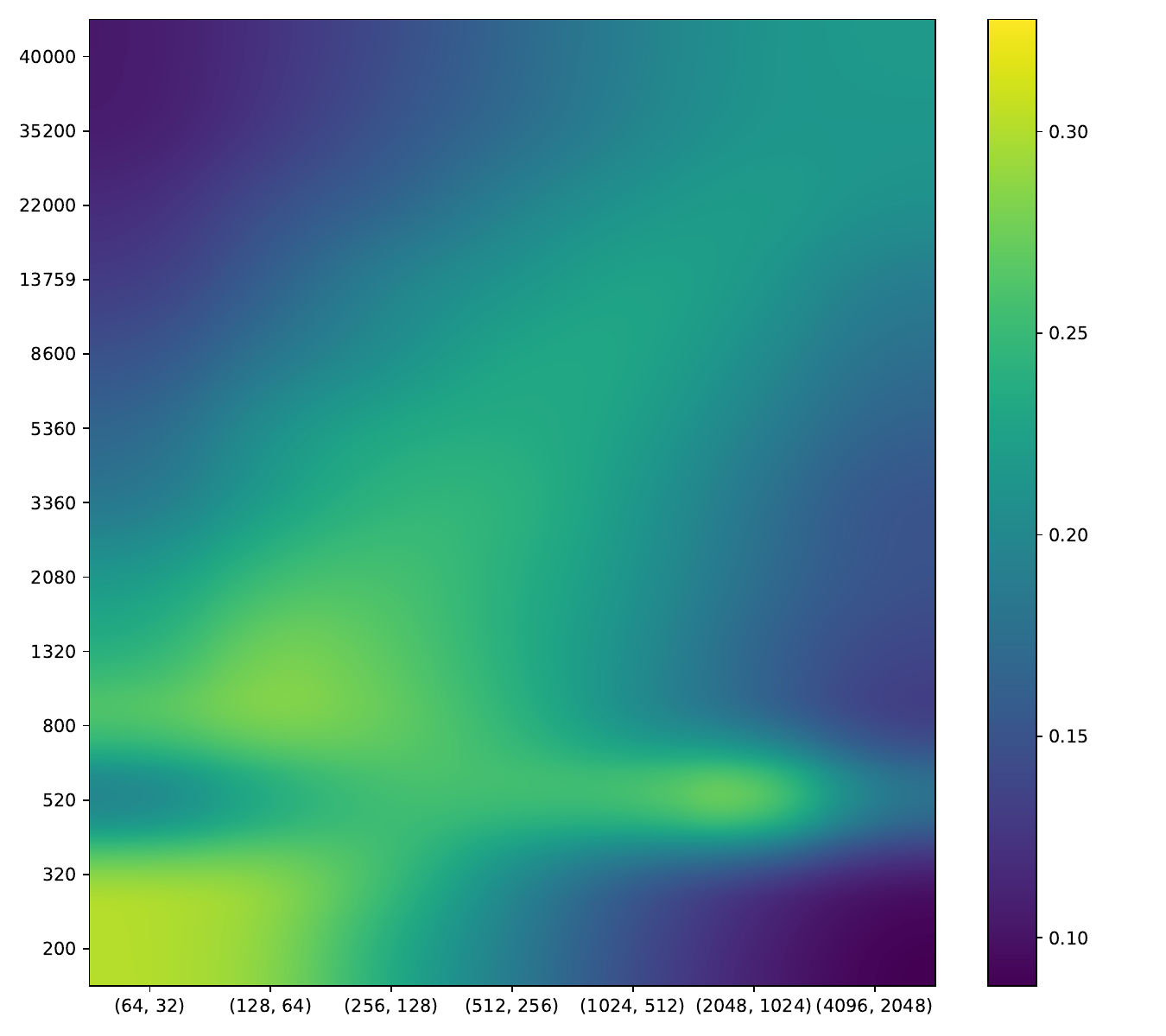}
         \caption{CIFAR10 - MC-Dropout}
         \label{fig:epis-uncer-id-cifar10-mc-dropout}
     \end{subfigure}
     \caption{Heatmaps of the normalized epistemic uncertainties on \textbf{ID} samples: the test sets of MNIST (\ref{fig:epis-uncer-id-mnist-ensemble},~\ref{fig:epis-uncer-id-mnist-mc-dropout}) and CIFAR10 (\ref{fig:epis-uncer-id-cifar10-ensemble},~\ref{fig:epis-uncer-id-cifar10-mc-dropout}). On the x-axis we have the number of neurons in the hidden layers and the number of samples used to train the models on the y-axis. Only the average of each test is reported.}
    \label{fig:epis-uncer-id}
\end{figure*}

%% accuracy: (mnist vs cifar10) and (ensemble vs MC-Dropout) 
\begin{figure*}[htbp]
     \centering
     \begin{subfigure}[b]{0.245\textwidth}
         \centering
         \includegraphics[width=\textwidth]{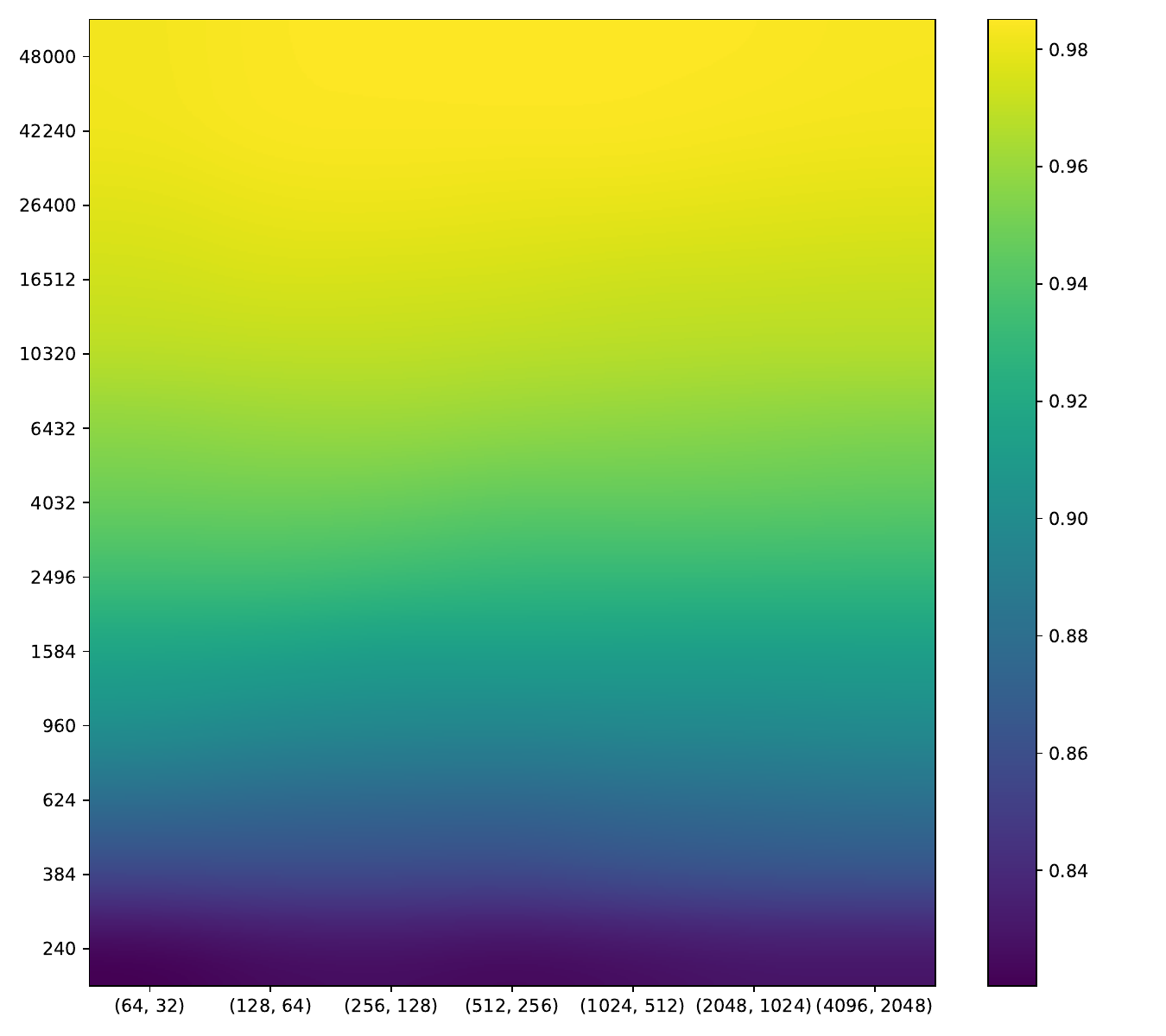}
         \caption{MNIST - Ensemble}
         \label{fig:accuracy-mnist-ensemble}
     \end{subfigure}
     \hfill
     \begin{subfigure}[b]{0.245\textwidth}
         \centering
         \includegraphics[width=\textwidth]{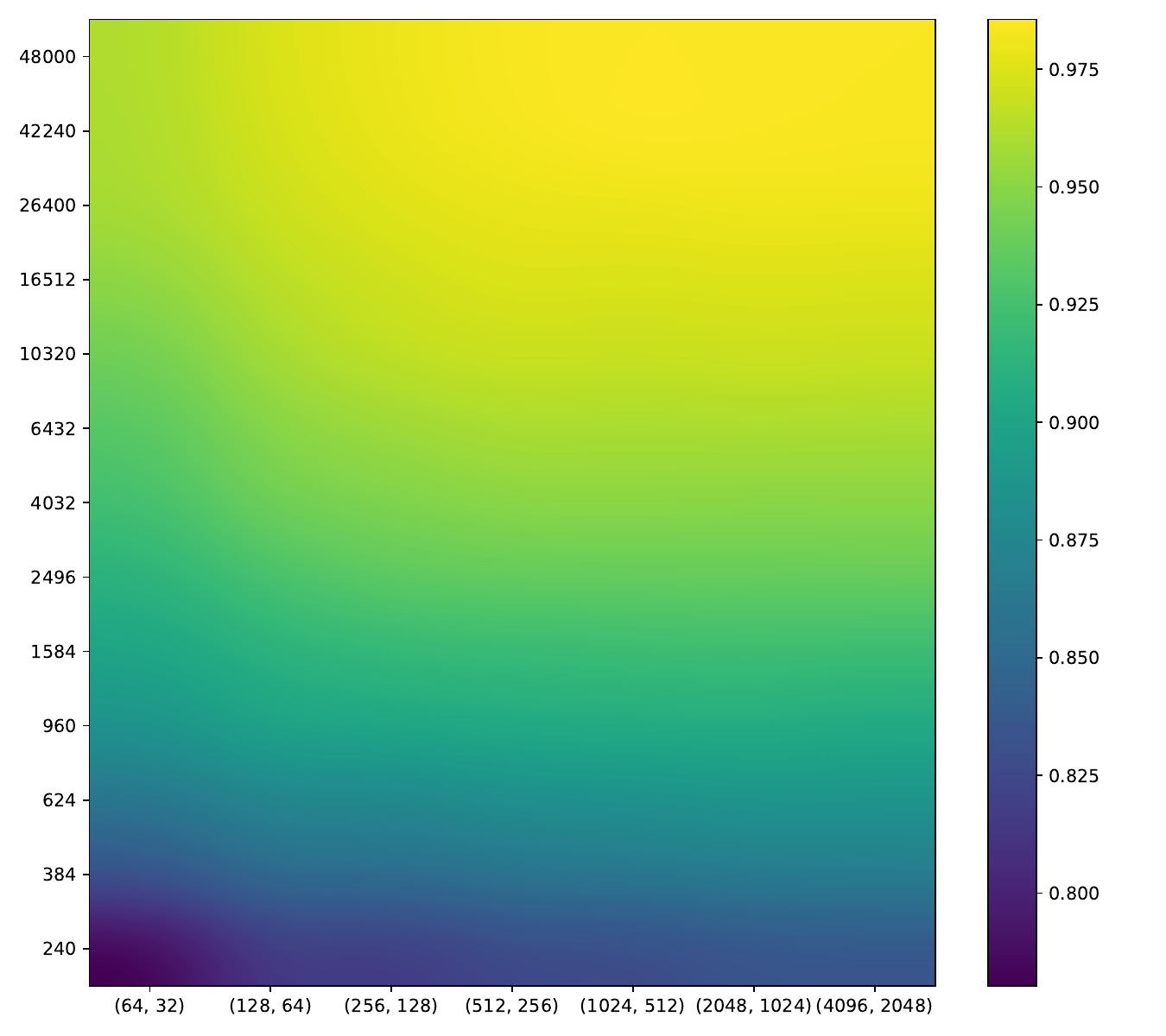}
         \caption{MNIST - MC-Dropout}
         \label{fig:accuracy-mnist-mc-dropout}
     \end{subfigure}
     \hfill
     \begin{subfigure}[b]{0.245\textwidth}
         \centering
         \includegraphics[width=\textwidth]{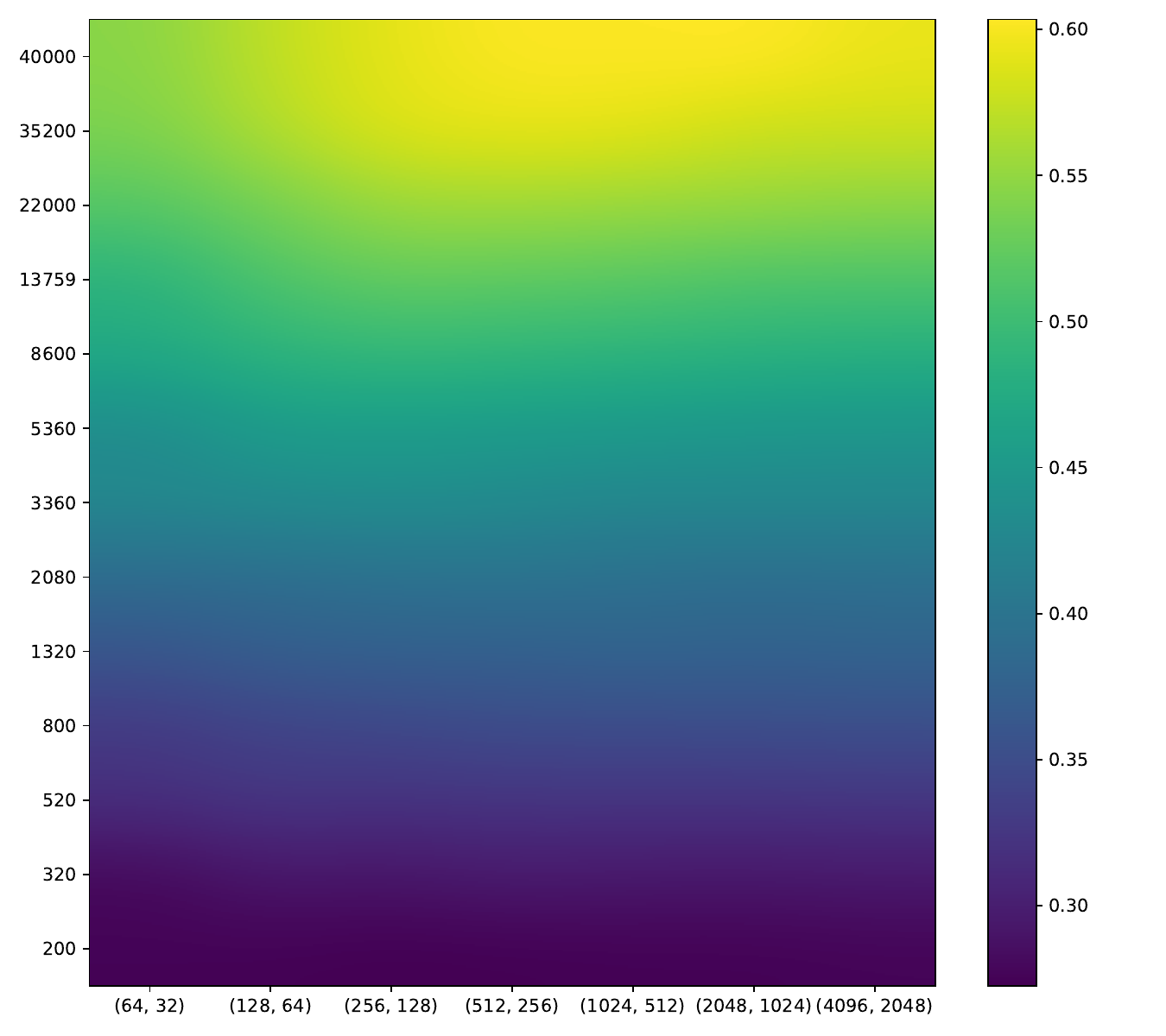}
         \caption{CIFAR10 - Ensemble}
         \label{fig:accuracy-cifar10-ensemble}
     \end{subfigure}
     \hfill
     \begin{subfigure}[b]{0.245\textwidth}
         \centering
         \includegraphics[width=\textwidth]{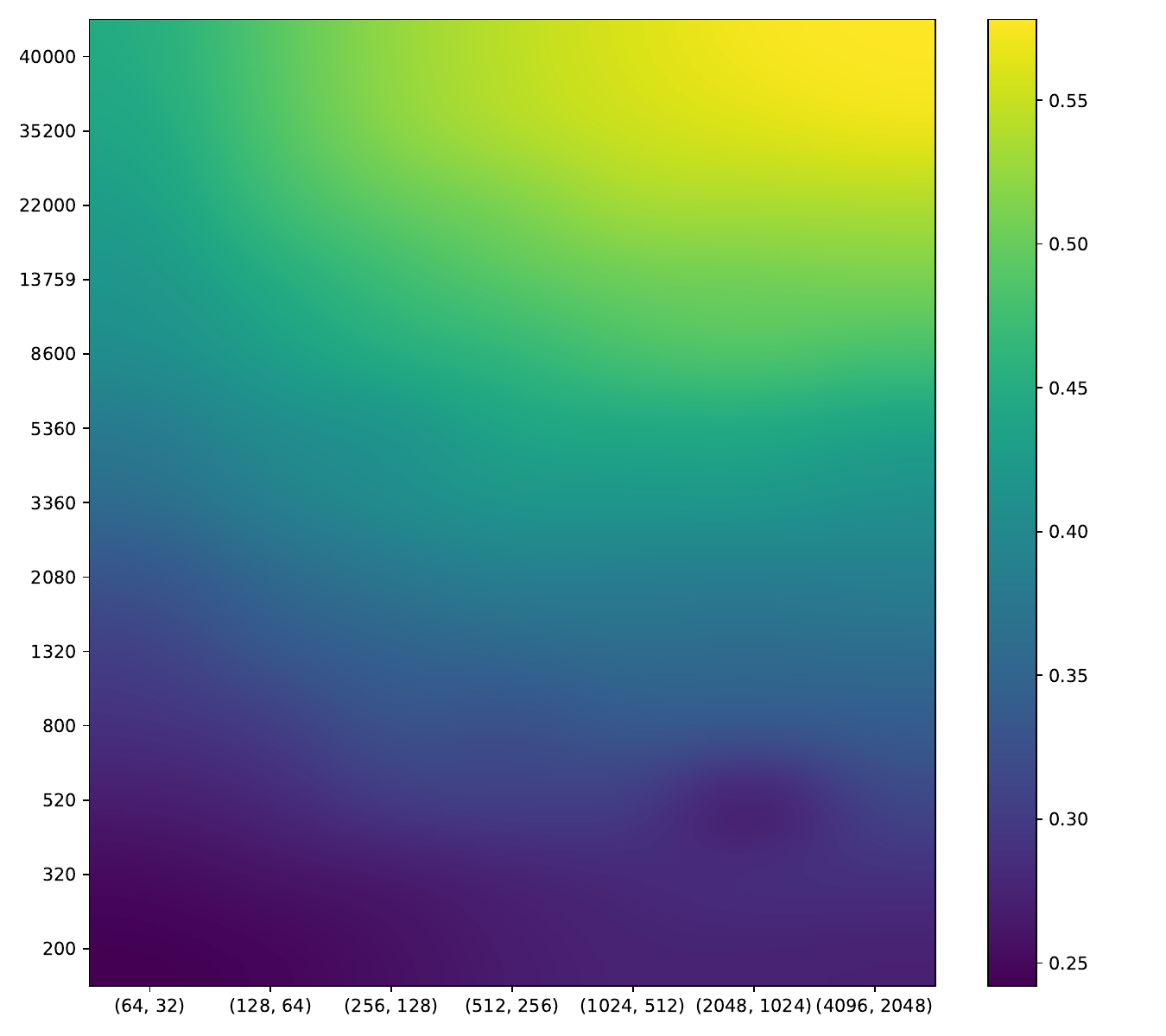}
         \caption{CIFAR10 - MC-Dropout}
         \label{fig:accuracy-cifar10-mc-dropout}
     \end{subfigure}
     \caption{Heatmaps of the models accuracy on \textbf{ID} samples: the test sets of MNIST (\ref{fig:accuracy-mnist-ensemble},~\ref{fig:accuracy-mnist-mc-dropout}) and CIFAR10 (\ref{fig:accuracy-cifar10-ensemble},~\ref{fig:accuracy-cifar10-mc-dropout}). On the x-axis we have the number of neurons in the hidden layers and the number of samples used to train the models on the y-axis.}
    \label{fig:accuracy-id}
\end{figure*}

\subsection{A motivating example\label{subsec:resnet18}}

We start by a motivating and prototypical example where we show that the epistemic uncertainty average $\bar{\mathcal{U}}_{epist.}$ is not a regular decreasing function of the number of training samples as we could expect. 
For this purpose, we have trained an ensemble of $K=10$ CNN ResNet18~\cite{he_deep_2015} models on subsets of CIFAR10 of varying length. The results are shown in Figure~\ref{fig:resnet18-cifar10}.

Considering in-distribution samples (\textbf{ID-all}), we see the epistemic uncertainty behaves as expected when the model is trained with a thousand images or more. However, below that threshold, uncertainty paradoxically increases with the number of training samples. We can take the analysis one step further by splitting \textbf{ID-all} into correctly classified examples (\textbf{ID-good}) and misclassified ones (\textbf{ID-mis}).
We see a clear separation of the corresponding box-plots for large training datasets. This confirms the proper functioning of the network, misclassified examples being on average more controversial. However we also notice that for small datasets, box-plots get aligned, and above all, less scattered. This suggests that epistemic uncertainty of networks trained on little data is non-informative.
%The epistemic uncertainty on the latter saturates for the models trained with more than $800$ images, while for the former we have the same trend as with the \textbf{ID-all} and the more data the models are trained on, the more the median of the epistemic uncertainty keeps decreasing.
We investigate this problem more in depth in the next subsection.
%Now that we have shown the "\emph{epistemic uncertainty hole}", we will see the effect of both the size of the model and the size of the training set.

\subsection{A two-dimensional analysis~\label{subsec:mlp}}

In this experiment, we will report the uncertainties as a function of both the number of parameters in the model and the size of the training set. We chose to work with Multilayer Perceptron (MLP) models rather than CNNs as controlling the size of MLPs is straightforward, by simply changing the number of neurons of hidden layers. As mentioned above, two types of BDL will be used: ensembles and MC-Dropout. For the former, we will use an ensemble of $10$ MLPs with $2$ hidden layers. For the latter, we keep the dropout layers~\cite{srivastava_dropout_nodate} stochastic in test mode and for a given input, we will compute $K$ outputs resulting from $K$ independent dropout samples. Both the hidden layers are followed with dropout layers with a rate of $0.5$. For the reported results, we used $K=30$. 
We also tested with $K=10$ and it yields the same results. 
We run the same experiment for both MNIST (Fig.~\ref{fig:epis-uncer-id-mnist-ensemble},~\ref{fig:epis-uncer-id-mnist-mc-dropout}) and CIFAR10 (Fig.~\ref{fig:epis-uncer-id-cifar10-ensemble},~\ref{fig:epis-uncer-id-cifar10-mc-dropout}) datasets. 
% Optional
The models are trained using Stochastic gradient descent (SGD) algorithm for $100$ epochs on MNIST and for $200$ on CIFAR10.

\subsubsection{MLP models trained on MNIST~\label{subsubsec:mnist}}
As one would expect, for a fixed model, the epistemic uncertainty does indeed decrease, when training it on more datapoints. However, when analyzing on a fixed train set, the epistemic uncertainty surprisingly decreases when increasing the number of parameters of the model. We also noticed that the aleatoric uncertainty is dominant in the predictive uncertainty term. This is true for both the ensemble models (Fig.~\ref{fig:epis-uncer-id-mnist-ensemble}) and MC-Dropout (Fig.~\ref{fig:epis-uncer-id-mnist-mc-dropout}).

\subsubsection{MLP models trained on CIFAR10~\label{subsubsec:cifar10}}

The behaviour of epistemic uncertainty gets even more paradoxal when considering a more complex dataset such as photos of CIFAR10.
This time, whether with ensembles (Fig.~\ref{fig:epis-uncer-id-cifar10-ensemble}) or MC dropout (Fig.~\ref{fig:epis-uncer-id-mnist-ensemble}), the epistemic uncertainty is neither a monotonic function for a fixed size of the training set nor for a fixed size of the model.
While for large models and few data (i.e. the bottom right corner of heatmaps) we expect a peak of uncertainty, we observe instead a depression, resulting in a clear diagonal ridge from bottom left to top right.
This unexpected depression is what we call the "epistemic uncertainty hole".

Also we note some differences on the results when using either ensembles or MC-Dropout. 
In particular a strange artefact, so far unexplained, appears with MC dropout (Fig.~\ref{fig:epis-uncer-id-cifar10-mc-dropout}) for medium-sized dataset (around 500 data samples). 

%For the ensembles, the epistemic uncertainty behaves similarly as the ResNet18 (Subsection~\ref{subsec:resnet18}) for the smaller MLP models. 
%In the presence of even more training samples, we might get the same trend for larger MLP models. In contrast, the epistemic uncertainty with MC-Dropout for models trained with large training set is higher for models with more parameters and hence as expected. However, with few examples, we notice similar results as with MNIST (\ref{subsubsec:mnist}).

%% epistemic uncertainties OOD: (mnist vs cifar10) and (ensemble vs MC-Dropout) 
\begin{figure*}[htbp]
     \centering
     \begin{subfigure}[b]{0.245\textwidth}
         \centering
         \includegraphics[width=\textwidth]{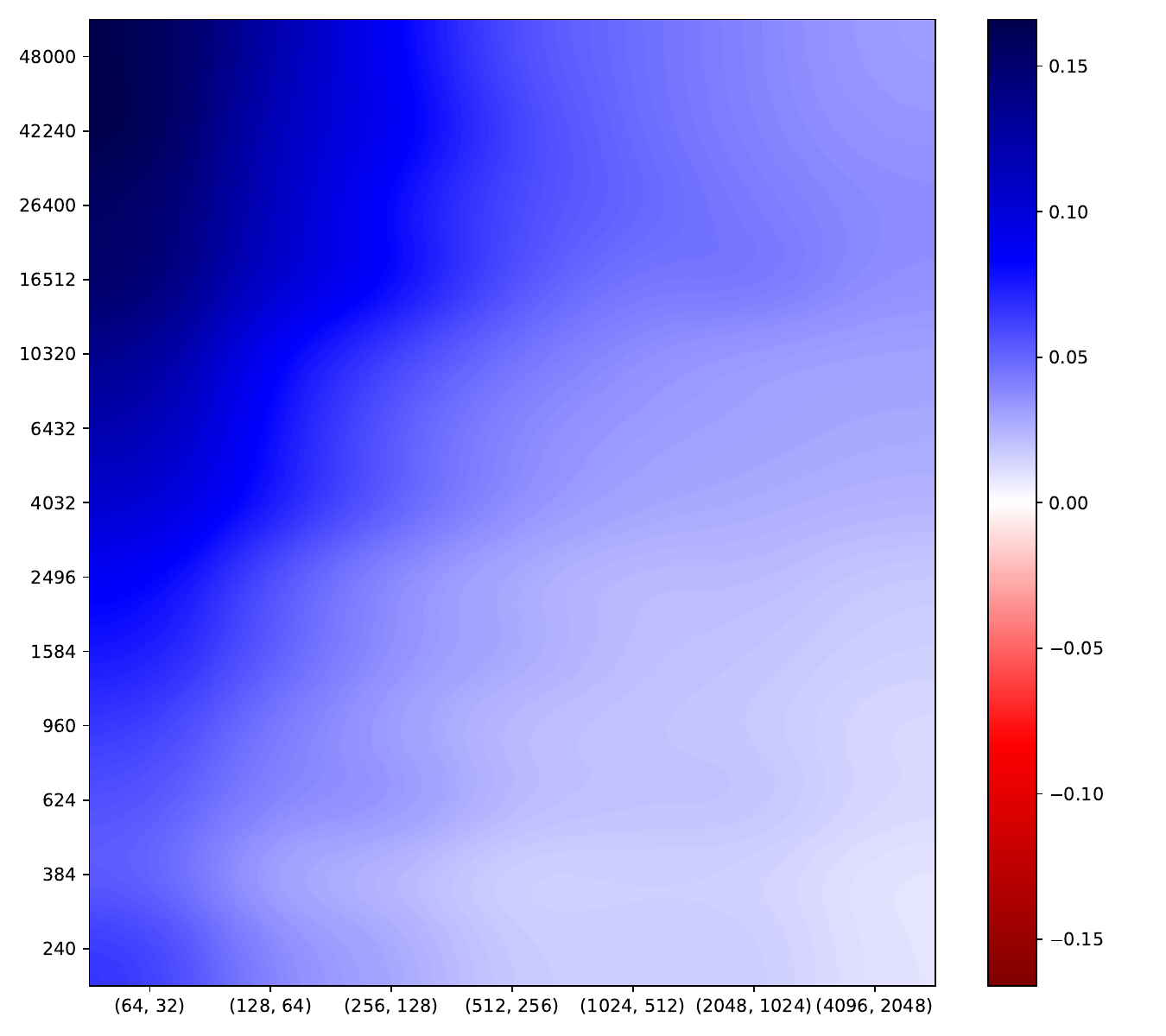}
         \caption{MNIST - Ensemble}
         \label{fig:epis-uncer-ood-mnist-ensemble}
     \end{subfigure}
     \hfill
     \begin{subfigure}[b]{0.245\textwidth}
         \centering
         \includegraphics[width=\textwidth]{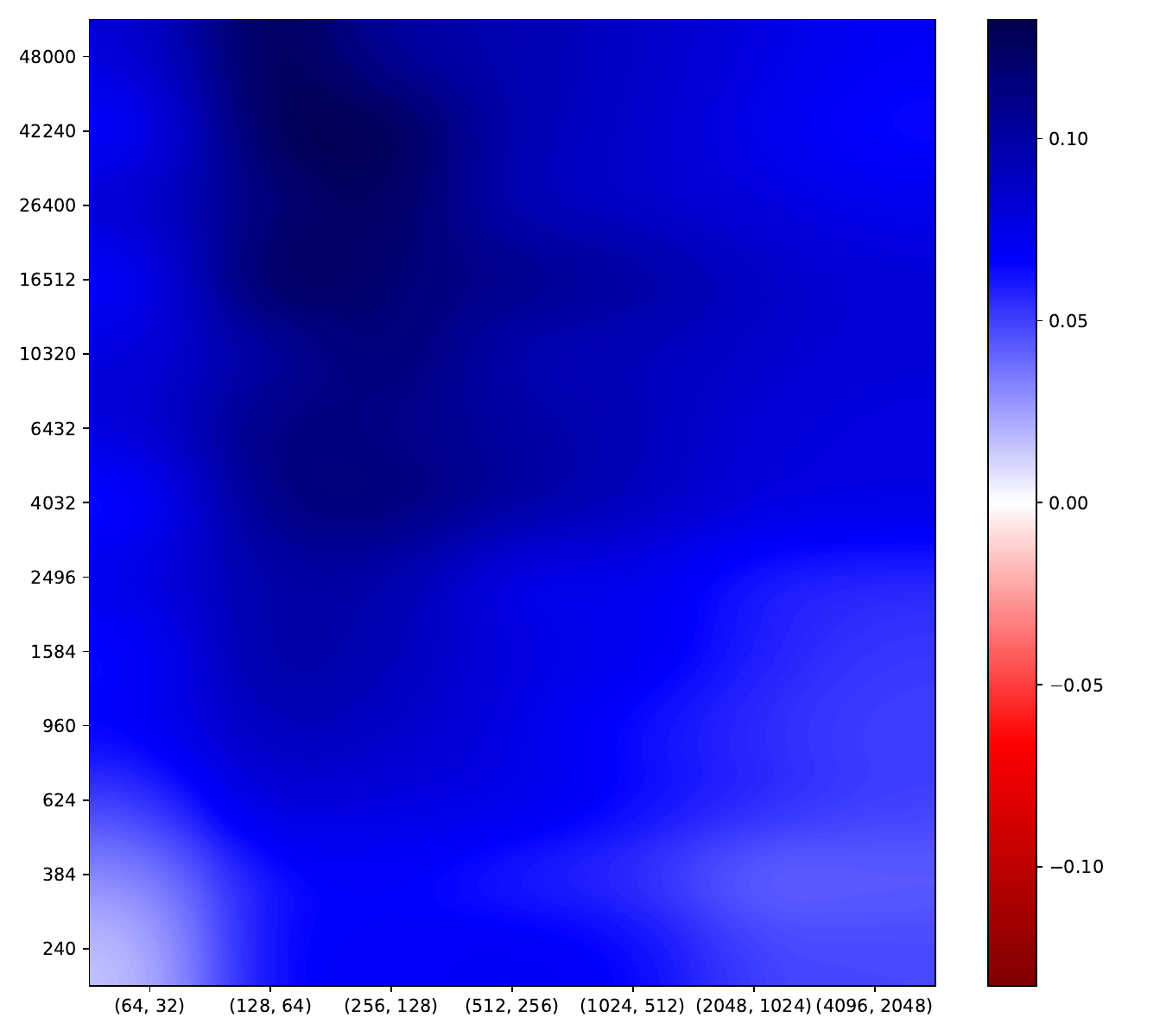}
         \caption{MNIST - MC-Dropout}
         \label{fig:epis-uncer-ood-mnist-mc-dropout}
     \end{subfigure}
     \hfill
     \begin{subfigure}[b]{0.245\textwidth}
         \centering
         \includegraphics[width=\textwidth]{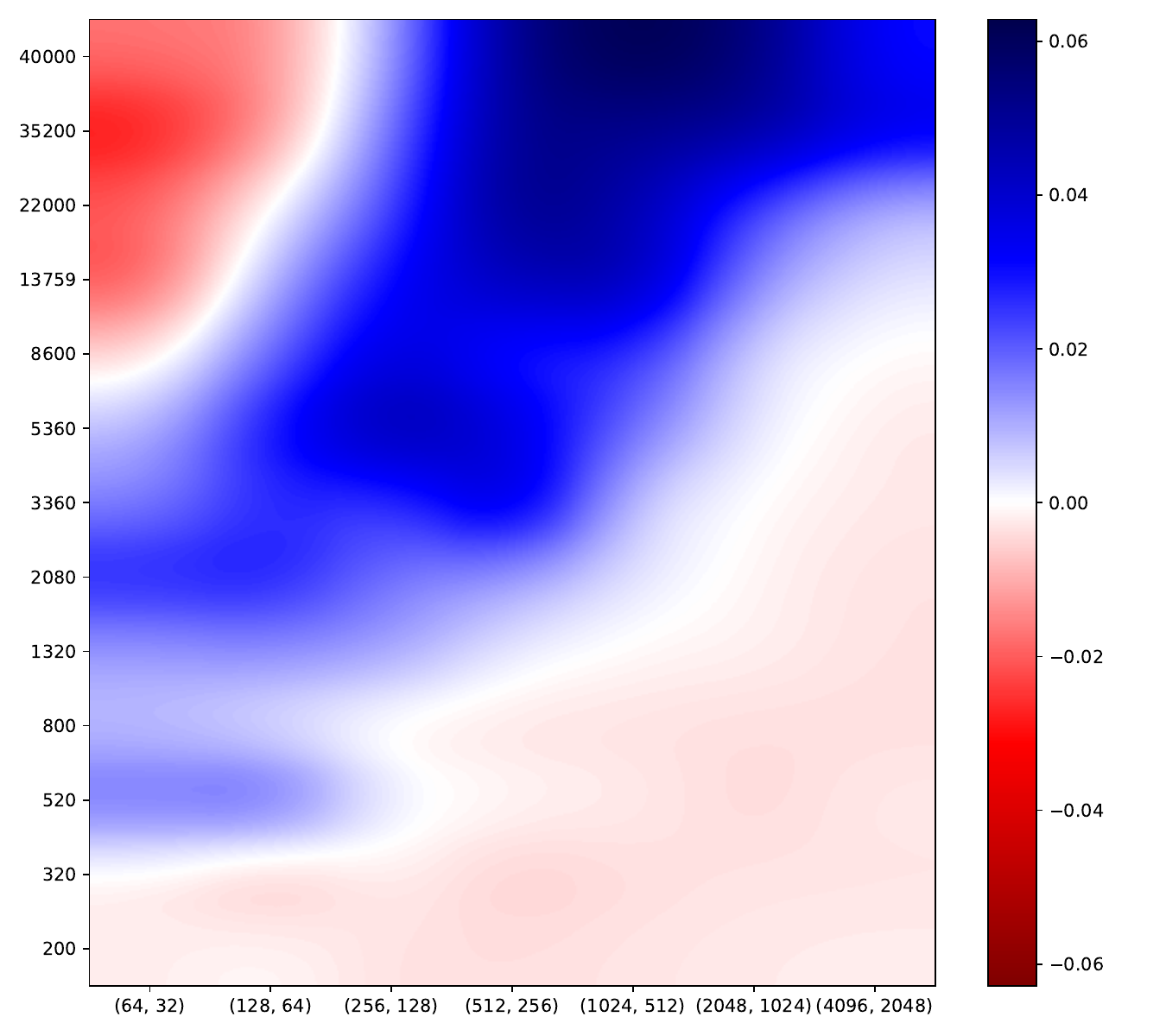}
         \caption{CIFAR10 - Ensemble}
         \label{fig:epis-uncer-ood-cifar10-ensemble}
     \end{subfigure}
     \hfill
     \begin{subfigure}[b]{0.245\textwidth}
         \centering
         \includegraphics[width=\textwidth]{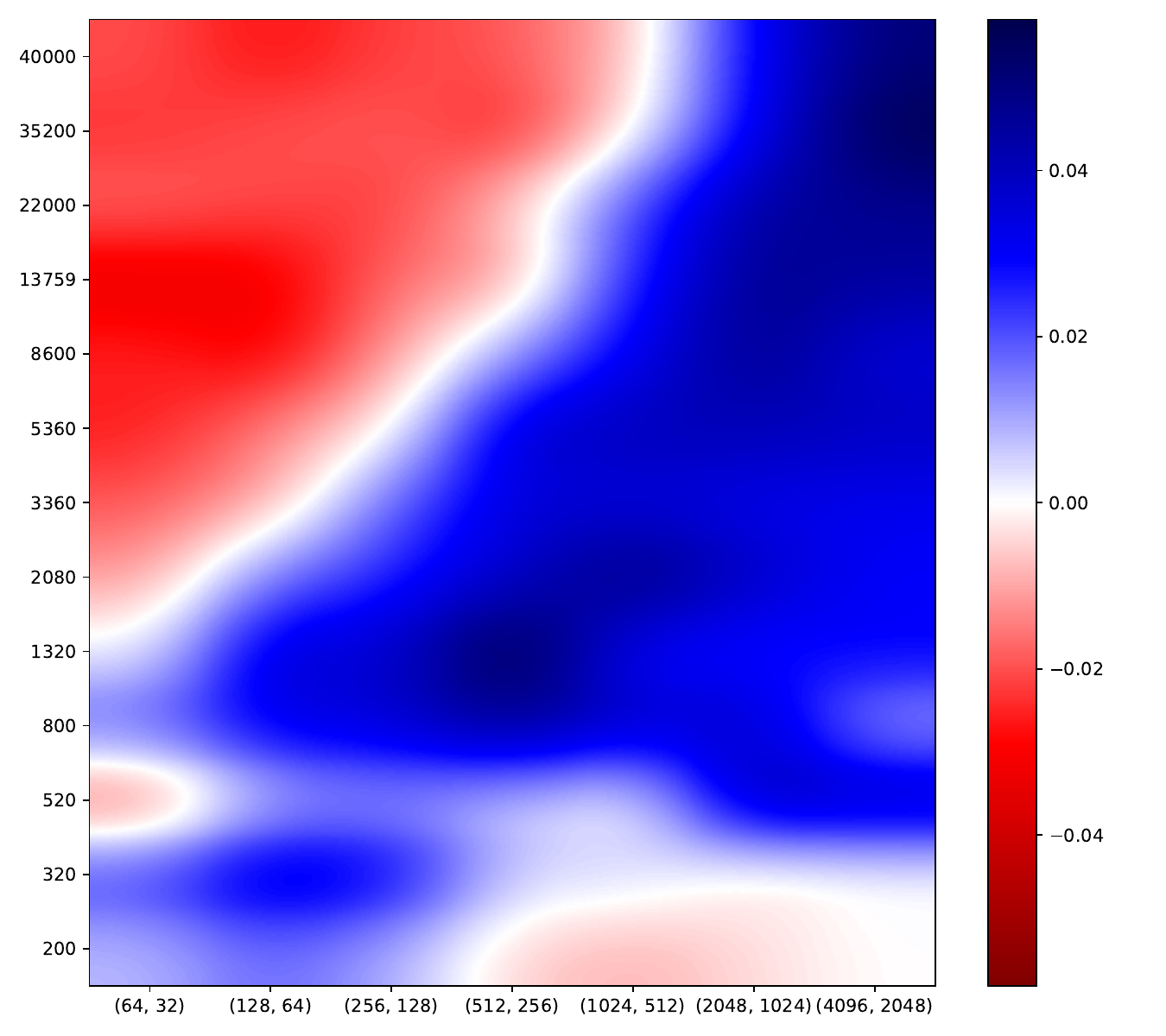}
         \caption{CIFAR10 - MC-Dropout}
         \label{fig:epis-uncer-ood-cifar10-mc-dropout}
     \end{subfigure}
     % : the test sets of FashionMNIST (\ref{fig:epis-uncer-ood-mnist-ensemble},~\ref{fig:epis-uncer-ood-mnist-mc-dropout}) and SVHN
     \caption{Difference between the normalized epistemic uncertainties on \textbf{OOD} samples and on \textbf{ID} samples. On the x-axis we have the number of neurons in the hidden layers and the number of samples used to train the models on the y-axis. Only the average of each test is reported.}
    \label{fig:epis-uncer-ood}
\end{figure*}

\begin{figure*}[htbp]
     \centering
     \begin{subfigure}[b]{0.245\textwidth}
         \centering
         \includegraphics[width=\textwidth]{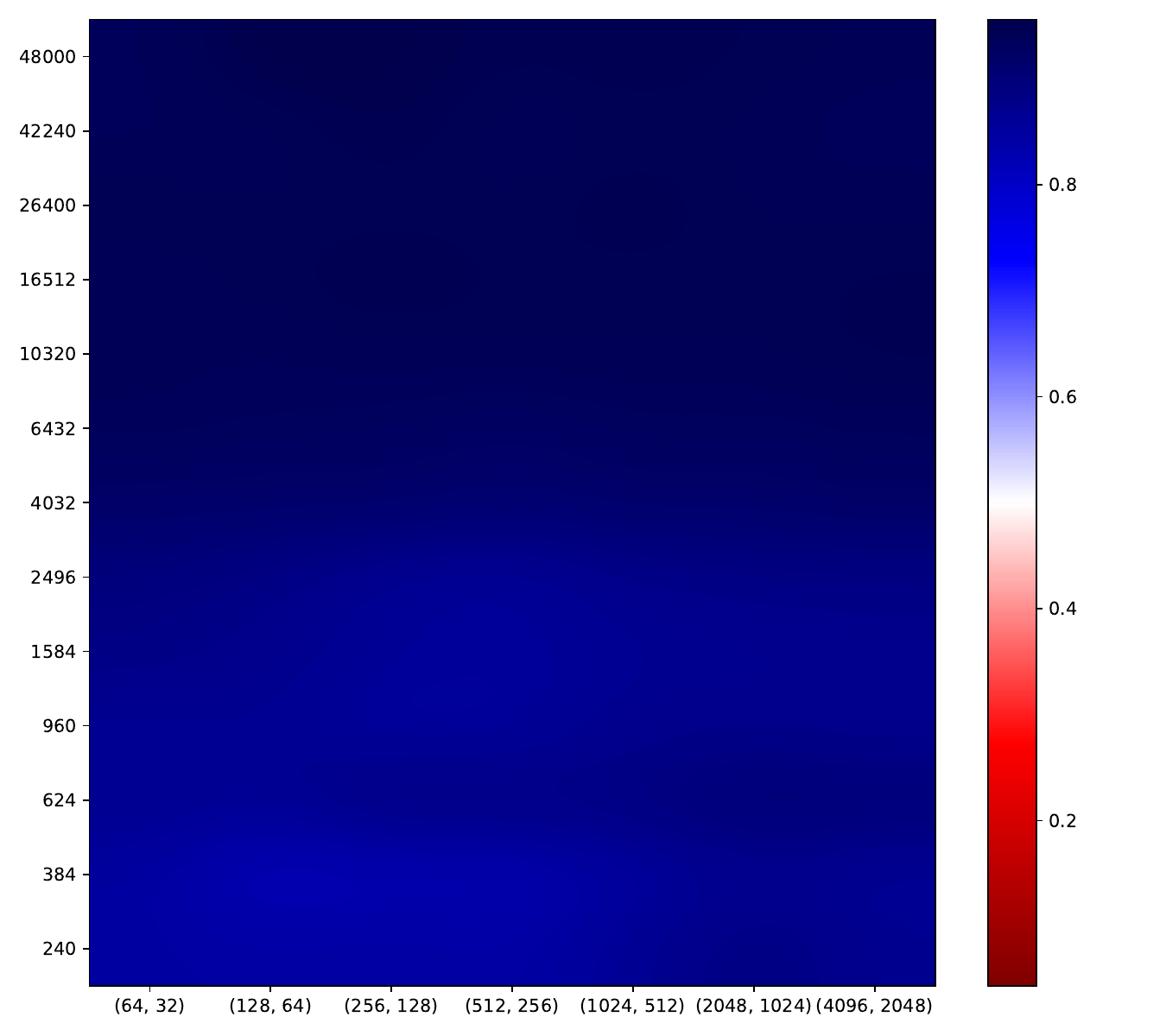}
         \caption{MNIST - Ensemble}
     \end{subfigure}
     \hfill
     \begin{subfigure}[b]{0.245\textwidth}
         \centering
         \includegraphics[width=\textwidth]{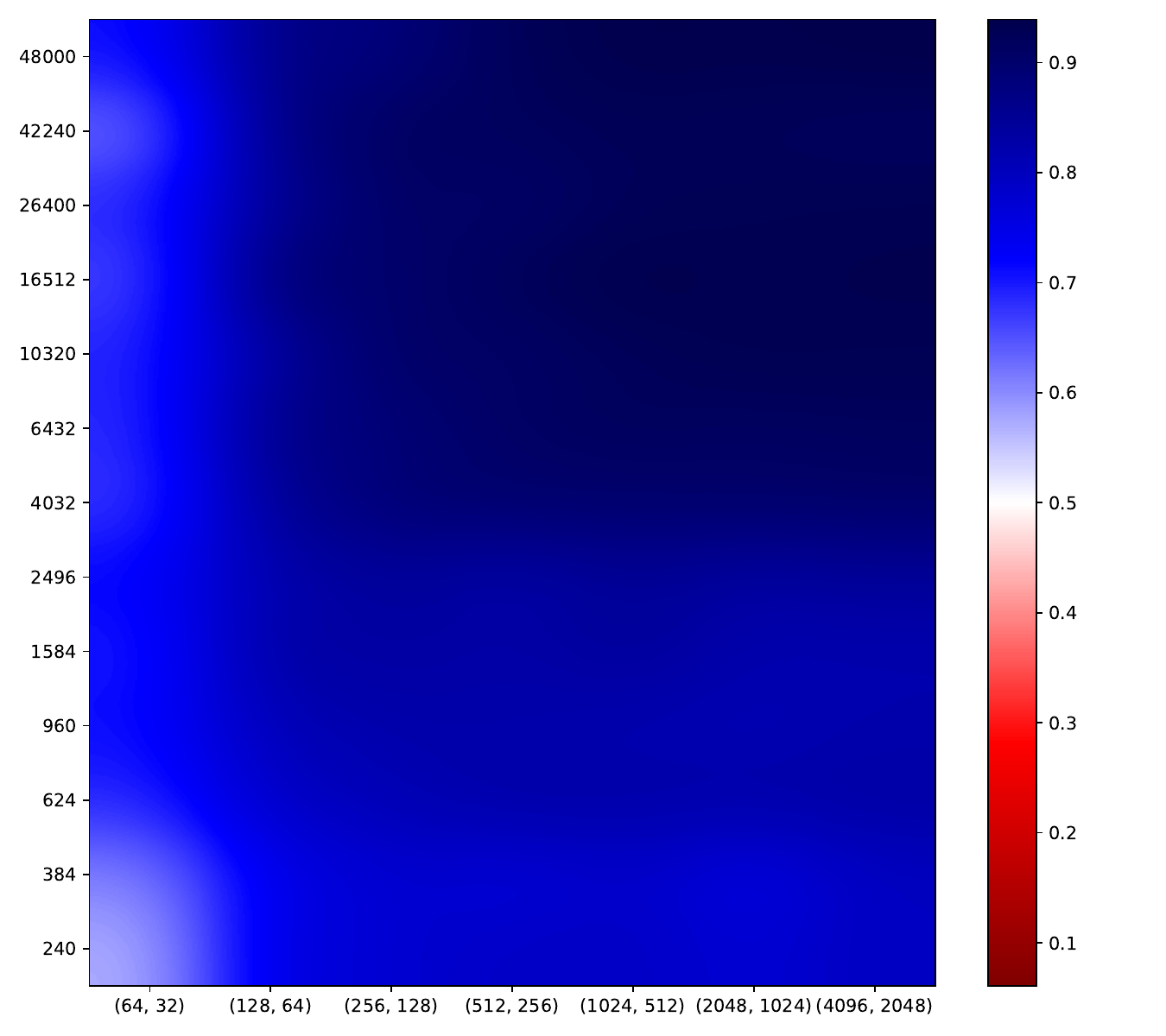}
         \caption{MNIST - MC-Dropout}
     \end{subfigure}
     \hfill
     \begin{subfigure}[b]{0.245\textwidth}
         \centering
         \includegraphics[width=\textwidth]{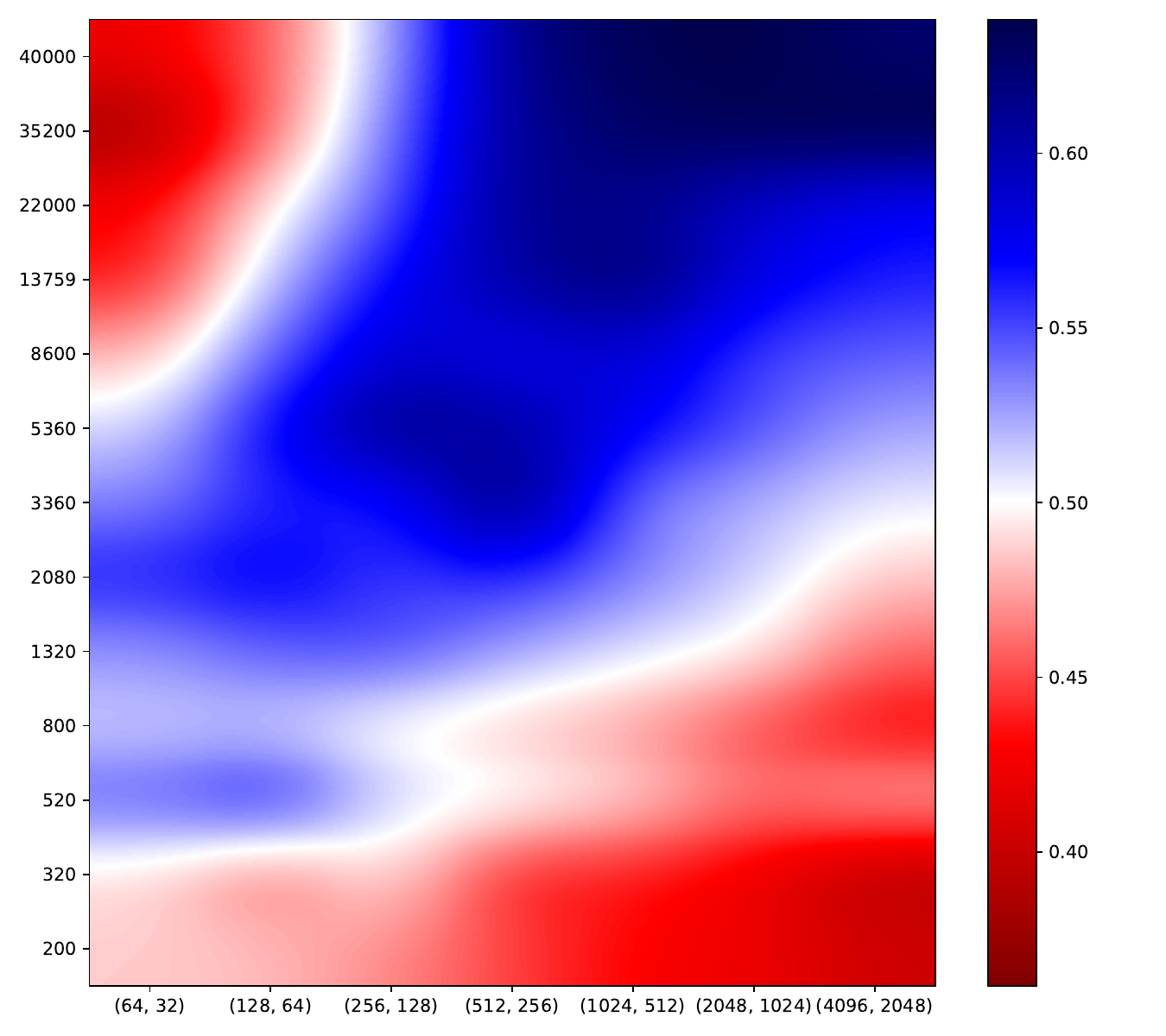}
         \caption{CIFAR10 - Ensemble}
     \end{subfigure}
     \hfill
     \begin{subfigure}[b]{0.245\textwidth}
         \centering
         \includegraphics[width=\textwidth]{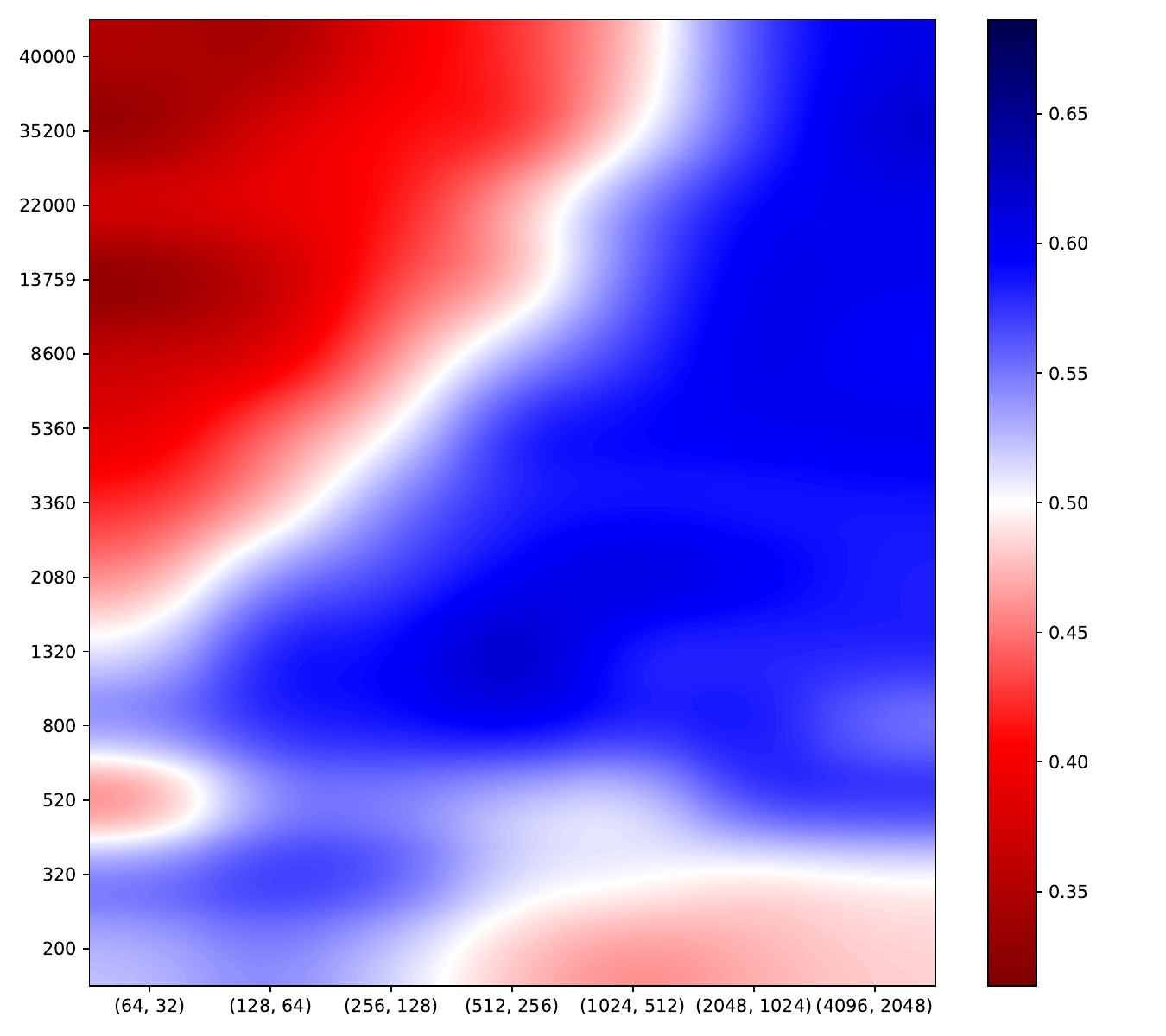}
         \caption{CIFAR10 - MC-Dropout}
     \end{subfigure}
     \caption{AUC score, centered at $0.5$, based on the normalized epistemic uncertainty: $0$ for \textbf{ID} and $1$ for \textbf{OOD}.}
     \label{fig:epis-uncer-auc}
\end{figure*}

\subsubsection{Performances and uncertainties}

It could be argued that we only considered the measures of epistemic uncertainty without checking the level of accuracy reached by the models. We can verify on Figure~\ref{fig:accuracy-id} that the networks work perfectly well from the sole point of view of accuracy: while CNNs or more sophisticated MLPs (e.g ResMLP~\cite{touvron2021}) can achieve better performance, the reached levels of accuracy are rather good for standard MLPs. Also for a fixed training set, the larger models are relatively more accurate than the smaller ones, with the ensemble models performing generally better than the MC-Dropout models (this is expected as the effective number of parameters of MC dropout should be multiplied by $1 - p_{dropout} = 0.5$  to make a fair comparison with ensembles). We do not observe any over-parametrization /overfitting effect as the accuracy does not drop for large models.
%MC dropout requires more parameters to perform well as expected (indeed) 
%Only the accuracy on the ensembles is less sensible to the model size than it is on MC-Dropout models. 
As a conclusion to this section, the accuracy of the model and its epistemic uncertainty are clearly two uncorrelated complementary pieces of information.

\section{Consequences on OOD detection~\label{sec:OOD}}

In this section, we assess the consequence of the epistemic uncertainty hole on the BDL application of OOD detection.
%One can use the measured uncertainties to detect the ambiguous or noisy input examples.
%In this section, we will focus on OOD examples.
As stated in Sect.~\ref{sec:BDL}, we expect that the average $\bar{\mathcal{U}}_{epist.}^{ood}$ of epistemic uncertainty computed on OOD samples be larger than the one $\bar{\mathcal{U}}_{epist.}^{id}$ computed on ID samples.
We verified this assumption by displaying on Fig.~\ref{fig:epis-uncer-ood} the gap $\Delta \bar{\mathcal{U}} = \bar{\mathcal{U}}_{epist.}^{ood} - \bar{\mathcal{U}}_{epist.}^{id}$ as a function of the model and training dataset sizes.
We used as OOD samples, the test sets of SVHN~\cite{netzer2011reading} and FashionMNIST~\cite{xiao2017fashionmnist} datasets for models trained on CIFAR10 and MNIST respectively.
For MNIST (Fig.~\ref{fig:epis-uncer-ood-mnist-ensemble} and Fig.~\ref{fig:epis-uncer-ood-mnist-mc-dropout}) results are globally consistent as the difference is positive everywhere (blue color).
However we already observe for ensembles (Fig.~\ref{fig:epis-uncer-ood-mnist-ensemble}) that the larger the model, the smaller the difference.
This counter-intuitive fact, that stems from the uncertainty hole, is less visible with MC dropout even if we can guess this phenomenon does not disappear but is simply pushed towards larger models (right edge of heatmap on Fig.~\ref{fig:epis-uncer-ood-mnist-mc-dropout}).

These issues are clearly amplified when testing on CIFAR10 (Fig.~\ref{fig:epis-uncer-ood-cifar10-ensemble} and Fig.~\ref{fig:epis-uncer-ood-cifar10-mc-dropout}).
In fact, the difference clearly appears negative (red color) in two opposite regions.
The one for large models and few data (bottom right of heatmaps) is consistent with the uncertainty hole.
%the distribution of epistemic uncertainty collapses and shrinks whether samples are ID or OOD.
%Coherent results are observed with the ResNet18's experiment when looking at\textbf{OOD} and \textbf{ID-all} boxplots on Fig.~\ref{fig:resnet18-cifar10}.
%
More surprising is the second regime for small models and many data (top left).
In that case, small networks trained on large datasets exhibit as expected, small epistemic uncertainties on ID examples.
The paradox is that OOD samples have even smaller epistemic uncertainties. We currently investigate the reason of this misbehavior, which is clearly not a spurious artefact.

We finally quantified the impact of these defects on the capacity of Bayesian neural networks to detect OOD samples.
To this end, we considered the binary classification problem consisting in separating OOD from ID samples given their levels of epistemic uncertainty.
We evaluated the AUC metric of such classifier. Results given on Figure~\ref{fig:epis-uncer-auc} appear fully consistent with the observations on the difference $\Delta \bar{\mathcal{U}}$. While OOD detection works well for large models with large data, the AUC gets lower than the AUC 0.5 of random classifiers in both previously discussed regions, illustrating the unexpected change of sign of $\Delta \bar{\mathcal{U}}$ in these areas.

% First, we continue the analysis of Subsection~\ref{subsec:resnet18}. Second,

% \subsection{OOD on ResNet18 models}
% As mentioned in Subsection~\ref{subsec:resnet18}, the misclassified examples have higher epistemic uncertainties compared to the correctly classified examples for the models trained on relatively larger training sets. Moreover, we have the same observation when comparing the OOD examples with the correctly classified examples. Generally speaking, the epistemic uncertainty of \textbf{OOD} and of \textbf{ID-mis} are consistently similar with the latter being higher (median-wise).

%\subsection{OOD on MLP models}
%We now test the models presented in Subsection~\ref{subsec:mlp} on OOD examples. The goal is to compare the average epistemic uncertainty on \textbf{ID} samples and on \textbf{OOD} samples as aforementioned. For the ensemble models trained on MNIST (Figure~\ref{fig:epis-uncer-ood-mnist-ensemble}), we see that the difference increases for a fixed model when training on more sample, thus OOD detection behaves as expected. For larger models, more data is used in order to make the distinction as small models do. The previous observation does not hold true the models trained on CIFAR10 (Figures~\ref{fig:epis-uncer-ood-cifar10-ensemble},~\ref{fig:epis-uncer-ood-cifar10-mc-dropout}): we have once again an epistemic uncertainty hole. 

\section{Conclusion~\label{sec:conclusion}}

In this article, different experiments consistently highlighted the existence of a ``hole'' of epistemic uncertainty as produced by Bayesian deep networks. This hole in turn entails negative consequences on real-world BDL applications, such as the detection of OOD samples. This immediately opens up two research perspectives.
The first one is to understand precisely the reasons of this hole and the factors influencing it. This clearly requires a simultaneous analysis of both epistemic and aleatoric uncertainties as they interact closely.
The second perspective is to design new corrective measures that will make Bayesian neural networks behave as expected with regard to epistemic uncertainty.
Only an effective response to this problem will make Bayesian neural networks a convincing solution, not only from a theoretical point of view but also and above all, from an application perspective.

%% references
\bibliography{biblio}
\bibliographystyle{plain}

\includeAppendix{
    \clearpage
    \newpage
    \onecolumn
    
    \newpage
    % \appendix
    
    \begin{appendices}
        \section{Implementation details~\label{sec:imp}}
        % PyTorch~\cite{NEURIPS2019_9015}
        All the models used for the experiments are implemented in python using PyTorch. The models trained on MNIST are trained for $100$ epochs and for $200$ those trained on CIFAR10 with a batch size of $128$. SGD optimized is used with a learning rate of $0.01$ and a momentum of $0.9$.
        
        In the upcoming sections, we will denote the different layers as following:
        \begin{itemize}
            \item $Lin(n)$: linear layer with $n$ being the dimension of the output 
            \item $Drop(p)$: Dropout layer with the probability $p$
            \item $ReLU()$: the rectified linear activation function
        \end{itemize}
        
        MNIST and CIFAR10 are classification tasks with 10 classes. Thus, the last layer will always be $Lin(10)$. In addition, to convert the logits to probability vectors, the \emph{softmax} function is used. 
        
        \subsection{The effect of the size of the training set~\label{subsec:imp-resnet18}}
        For the experiments of Subsection~\ref{subsec:resnet18}, we use the model initialized from scratch provided by \emph{torchvision} package. The only change applied to this model is that the fully connected layer is replaced with an MLP: $Lin(512)-ReLU()-Lin(256)-ReLU()-Lin(10)$.
        
        \subsection{A two-dimensional analysis~\label{subsec:imp-mlp}}
        In Subsection~\ref{subsec:mlp}, we change the number of neurons in the two hidden layers ($h_1$ and $h_2$) of the MLP. In the case of an ensemble model, the MLP is as follows: $Lin(h_1)-Drop(0.1)-ReLU()-Lin(h_2)-Drop(0.1)-ReLU()-Lin(10)$. The same model is used for MC-Dropout with the only difference being the dropout rates set at $0.5$ for the highest variance: $Lin(h_1)-Drop(0.5)-ReLU()-Lin(h_2)-Drop(0.5)-ReLU()-Lin(10)$
        
        \section{Datasets}
        Here a reminder of the length of each test set of the datasets used in this paper:
        
        \begin{center}
            \begin{tabular}{|l|c|c|c|c|}
                \hline
                & MNIST & CIFAR10 & FashionMNIST & SVHN \\
                \hline
                number of examples: & 10000 & 10000 & 10000 & 26032 \\
                \hline
            \end{tabular}
        \end{center}
        
    \end{appendices}
}    

%% ONLY in detailled version !!!!!
\inDetailledVersion{
    \newpage
    the remaining pages are only for the detailed version, they won't be on the final version ...
    \section*{MC-Dropout - rates $0.5$-$0.5$}
    \begin{figure*}[h!]
         \centering
         \begin{subfigure}[b]{0.245\textwidth}
             \centering
             \includegraphics[width=\textwidth]{figures/MC-Dropout/mlp/mnist/30-forward-passes/mean_mutual_information_id_loader.pdf}
             \caption{MNIST - ID}
         \end{subfigure}
         \hfill
         \begin{subfigure}[b]{0.245\textwidth}
             \centering
             \includegraphics[width=\textwidth]{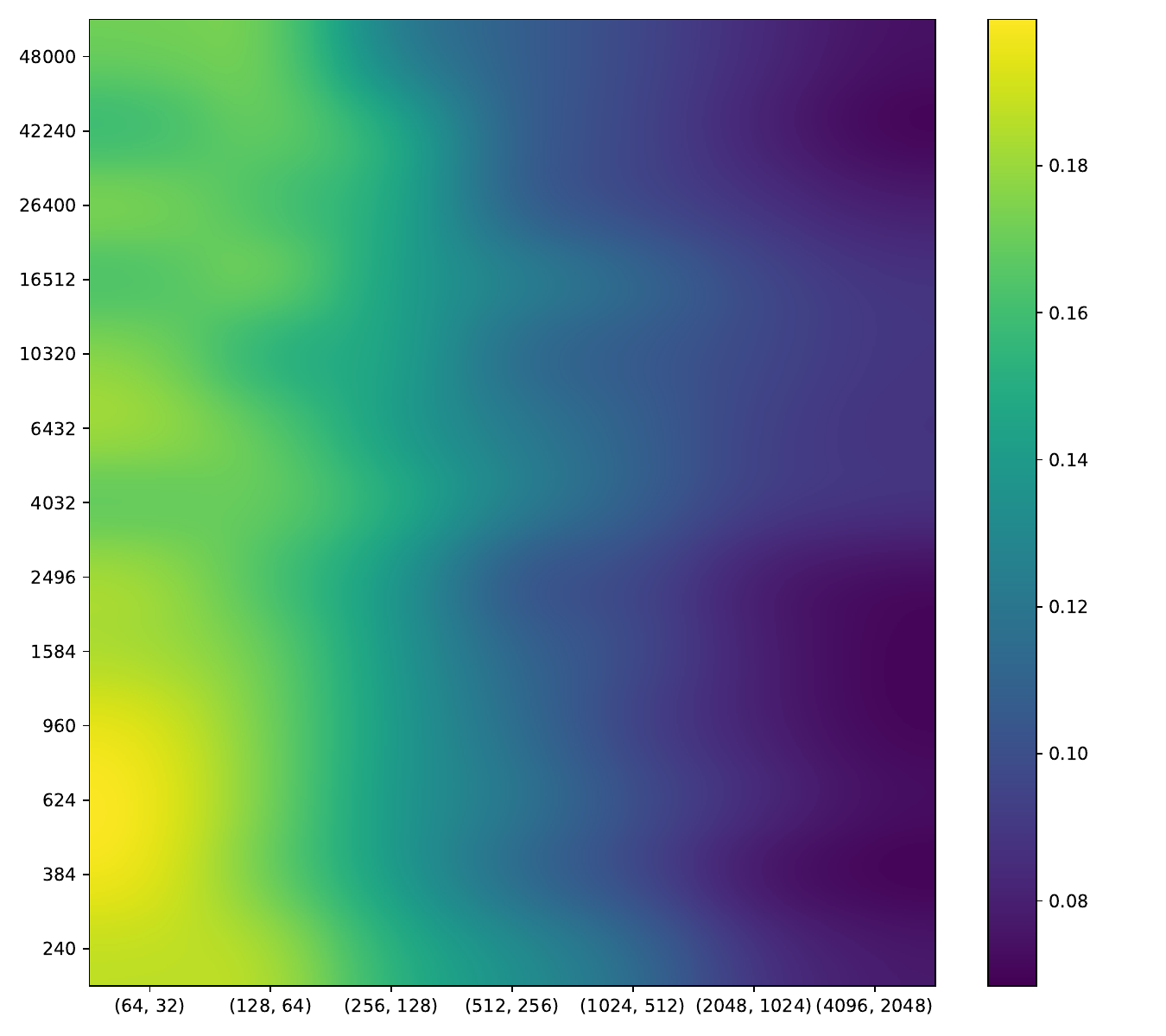}
             \caption{MNIST - OOD}
         \end{subfigure}
         \hfill
         \begin{subfigure}[b]{0.245\textwidth}
             \centering
             \includegraphics[width=\textwidth]{figures/MC-Dropout/mlp/cifar10-200epochs/30-forward-passes/mean_mutual_information_id_loader.pdf}
             \caption{CIFAR10 - ID}
         \end{subfigure}
         \hfill
         \begin{subfigure}[b]{0.245\textwidth}
             \centering
             \includegraphics[width=\textwidth]{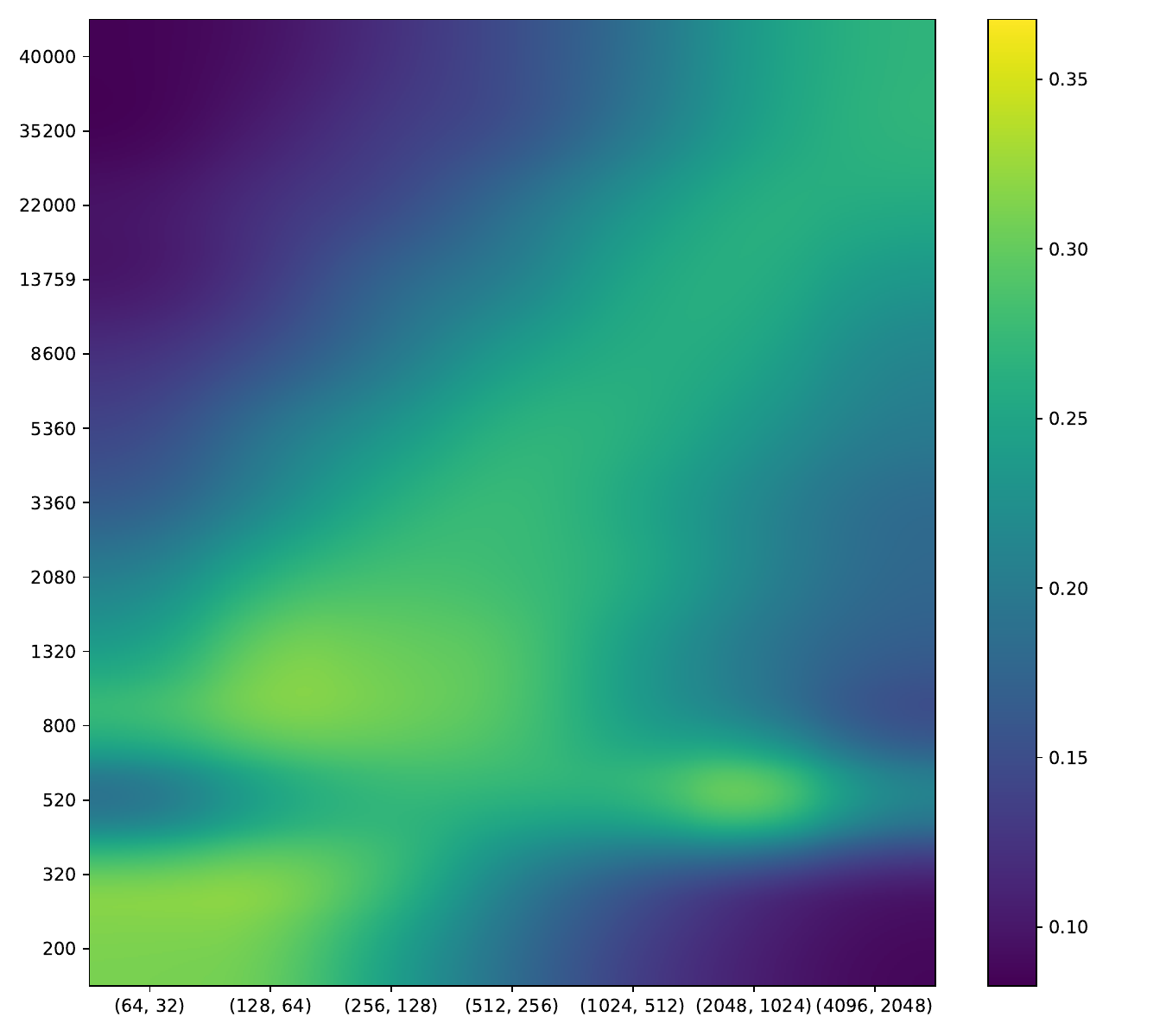}
             \caption{CIFAR10 - OOD}
         \end{subfigure}
         \caption{[Epistemic Uncertainty] MC-Dropout rates $0.5$-$0.5$}
    \end{figure*}
    
    \section*{MC-Dropout - rates $0.5$-$0.1$}
    \begin{figure*}[h!]
         \centering
         \begin{subfigure}[b]{0.245\textwidth}
             \centering
             \includegraphics[width=\textwidth]{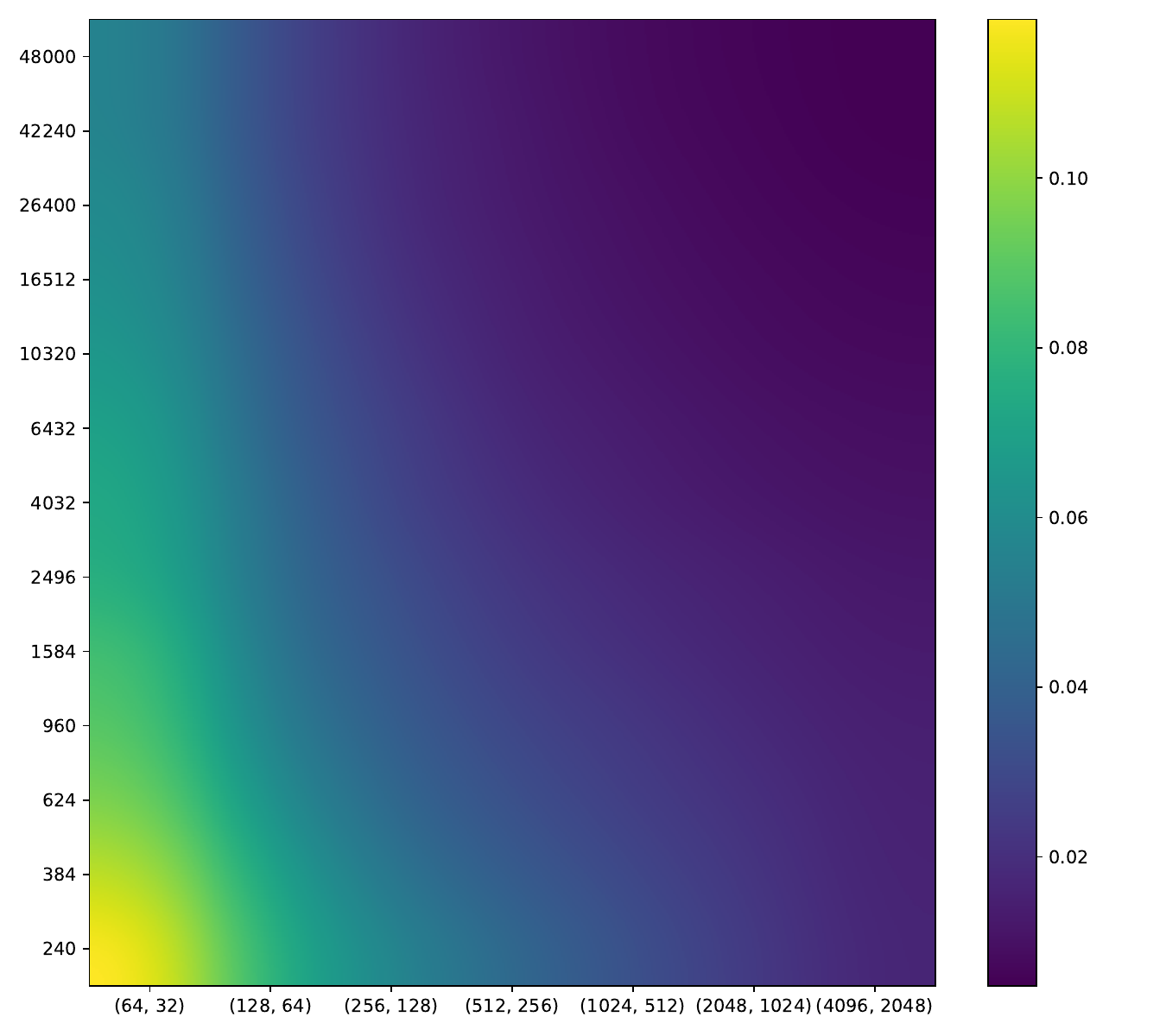}
             \caption{MNIST - ID}
         \end{subfigure}
         \hfill
         \begin{subfigure}[b]{0.245\textwidth}
             \centering
             \includegraphics[width=\textwidth]{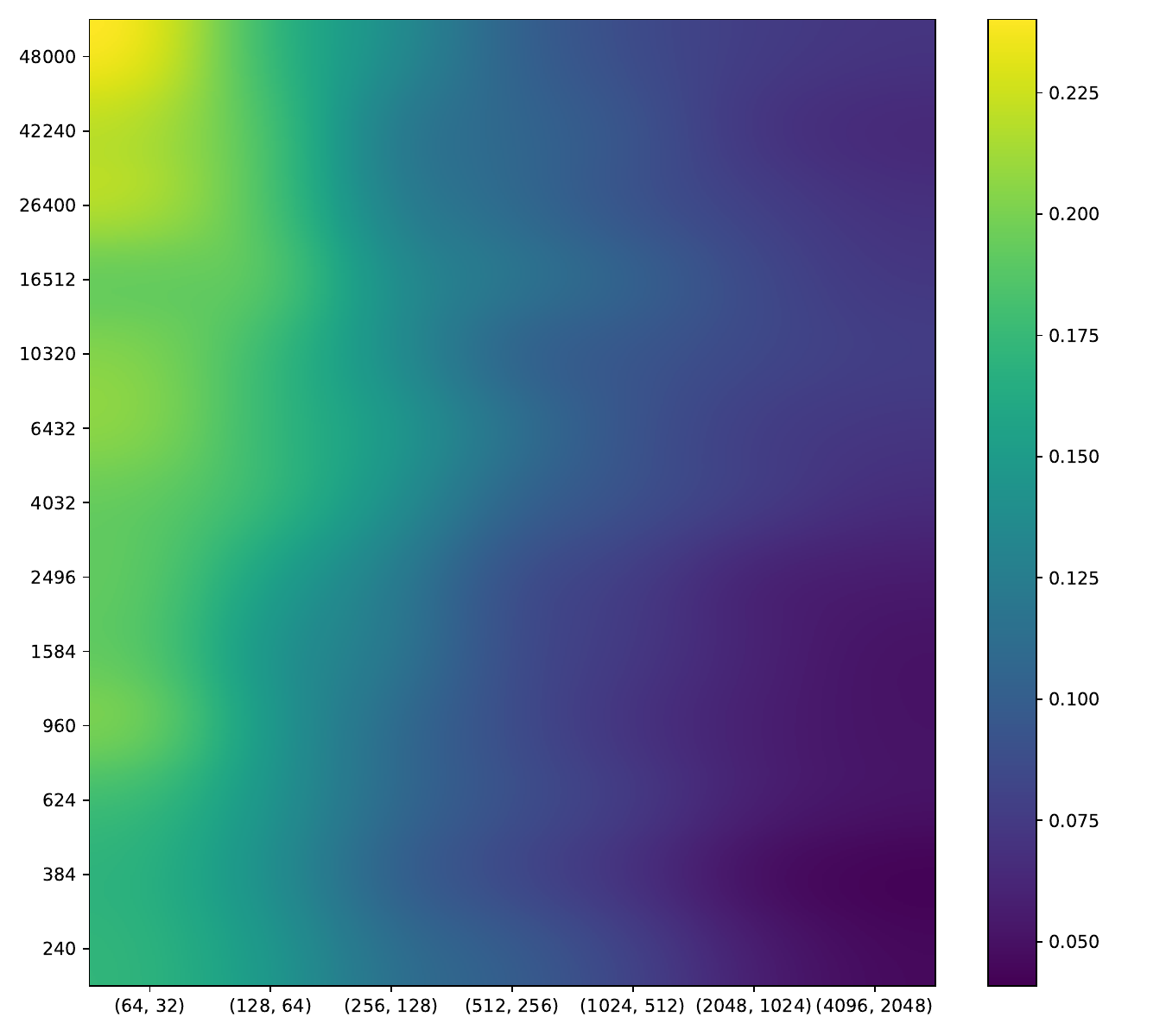}
             \caption{MNIST - OOD}
         \end{subfigure}
         \hfill
         \begin{subfigure}[b]{0.245\textwidth}
             \centering
             \includegraphics[width=\textwidth]{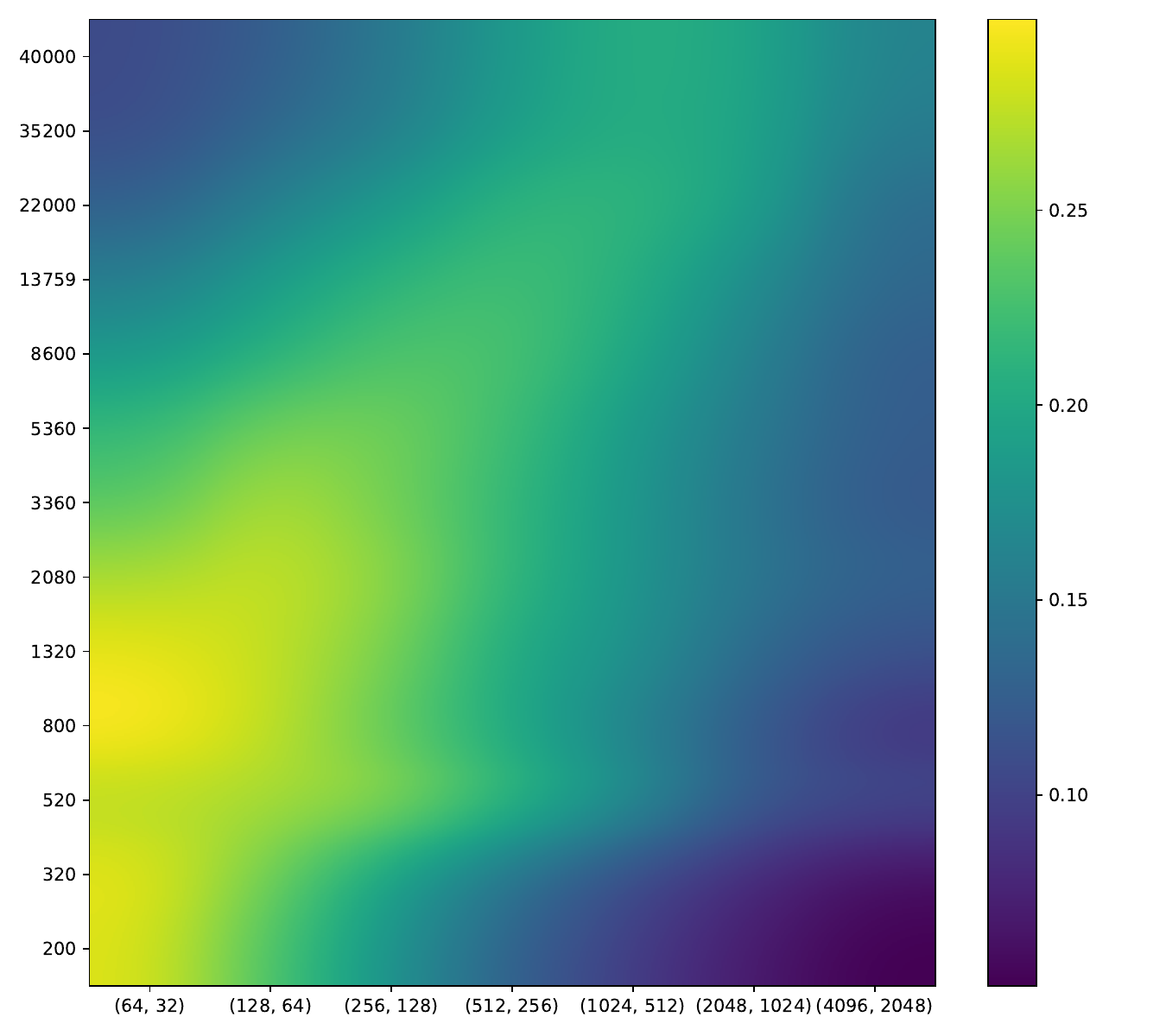}
             \caption{CIFAR10 - ID}
         \end{subfigure}
         \hfill
         \begin{subfigure}[b]{0.245\textwidth}
             \centering
             \includegraphics[width=\textwidth]{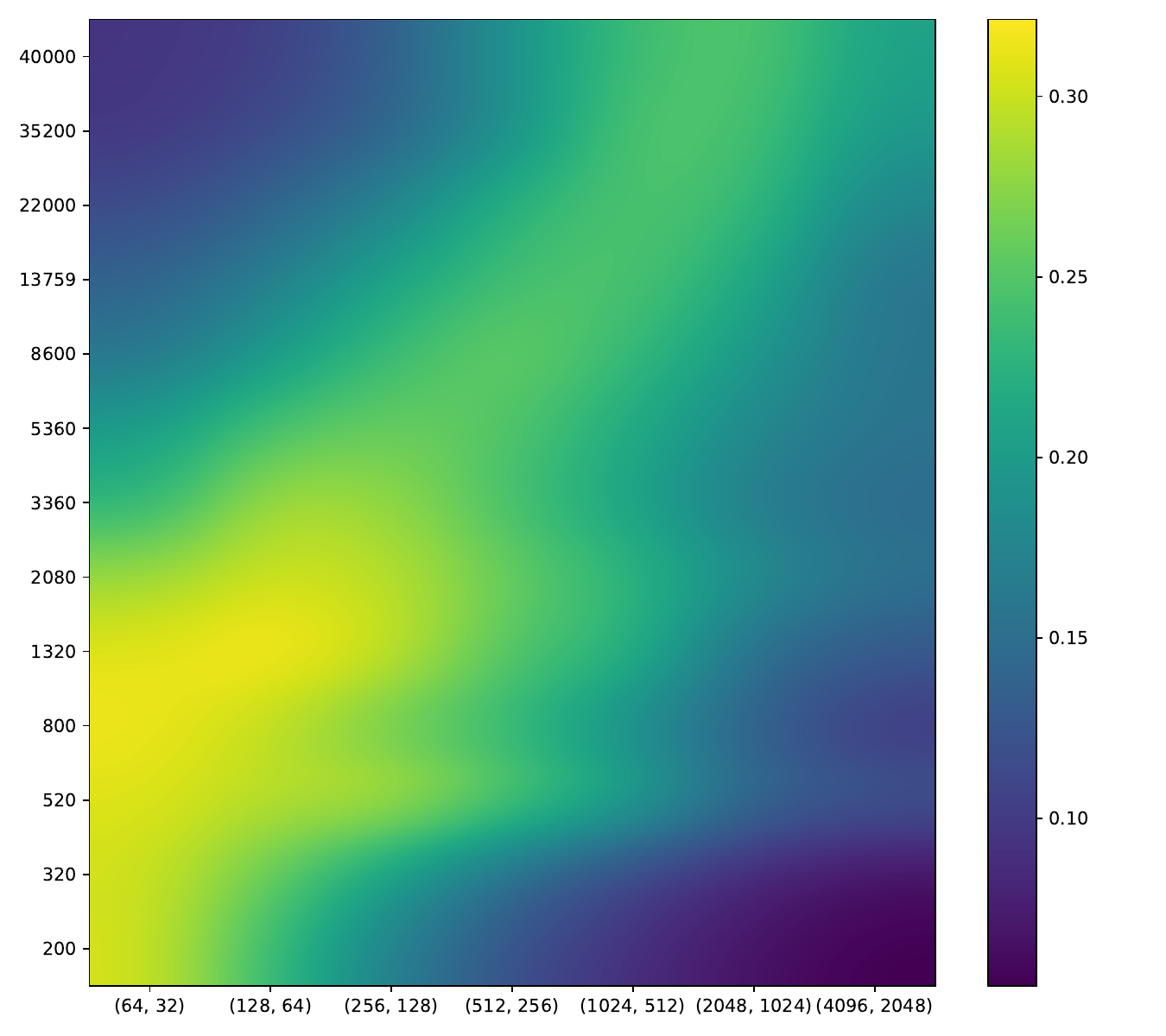}
             \caption{CIFAR10 - OOD}
         \end{subfigure}
         \caption{[Epistemic Uncertainty] MC-Dropout rates $0.5$-$0.1$}
    \end{figure*}

    \section*{Ensemble}
    \begin{figure*}[h!]
         \centering
         \begin{subfigure}[b]{0.245\textwidth}
             \centering
             \includegraphics[width=\textwidth]{figures/ensemble/mlp/mnist/mean_mutual_information_id_loader.pdf}
             \caption{MNIST - ID}
         \end{subfigure}
         \hfill
         \begin{subfigure}[b]{0.245\textwidth}
             \centering
             \includegraphics[width=\textwidth]{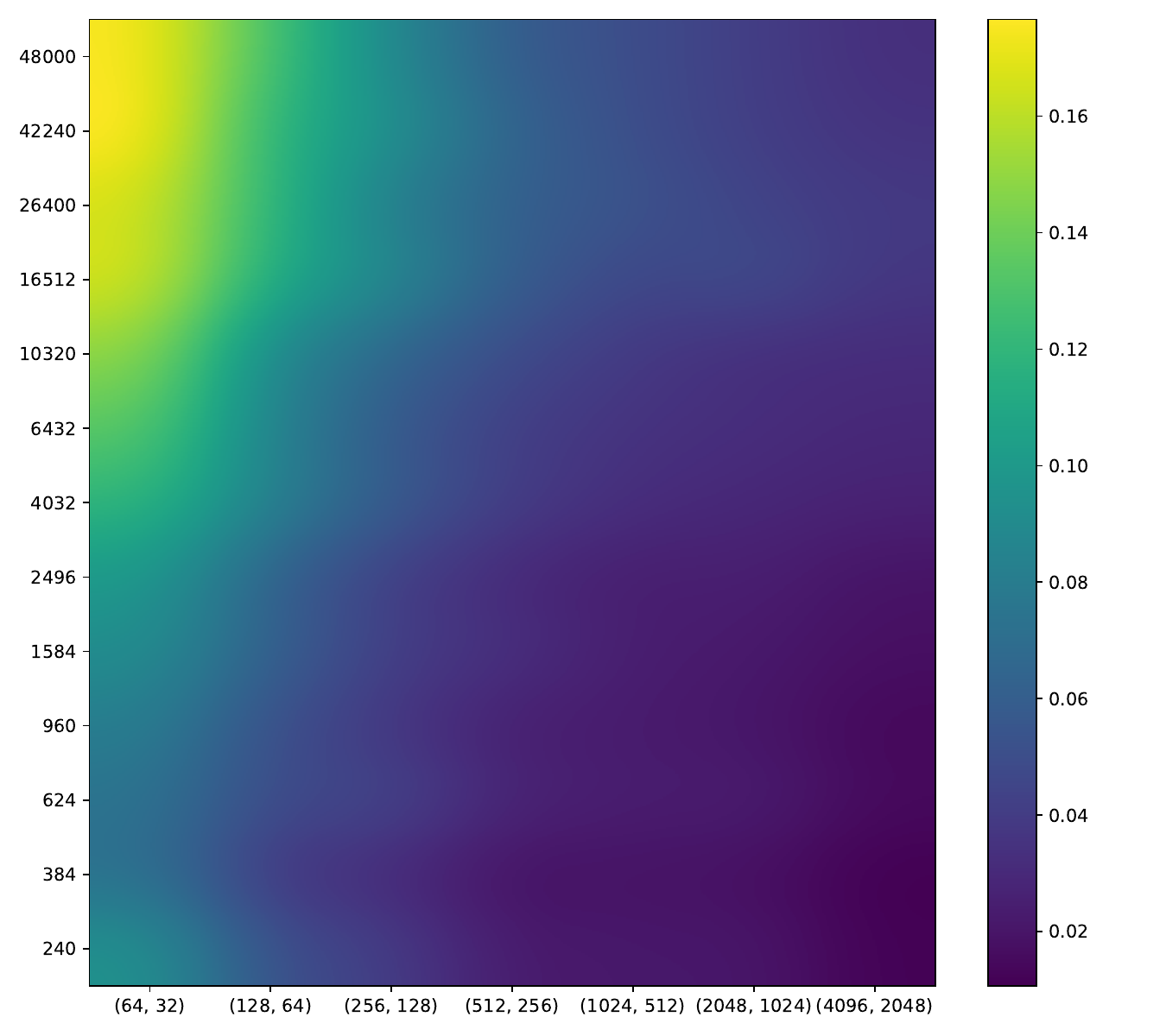}
             \caption{MNIST - OOD}
         \end{subfigure}
         \hfill
         \begin{subfigure}[b]{0.245\textwidth}
             \centering
             \includegraphics[width=\textwidth]{figures/ensemble/mlp/cifar10-200epochs/mean_mutual_information_id_loader.pdf}
             \caption{CIFAR10 - ID}
         \end{subfigure}
         \hfill
         \begin{subfigure}[b]{0.245\textwidth}
             \centering
             \includegraphics[width=\textwidth]{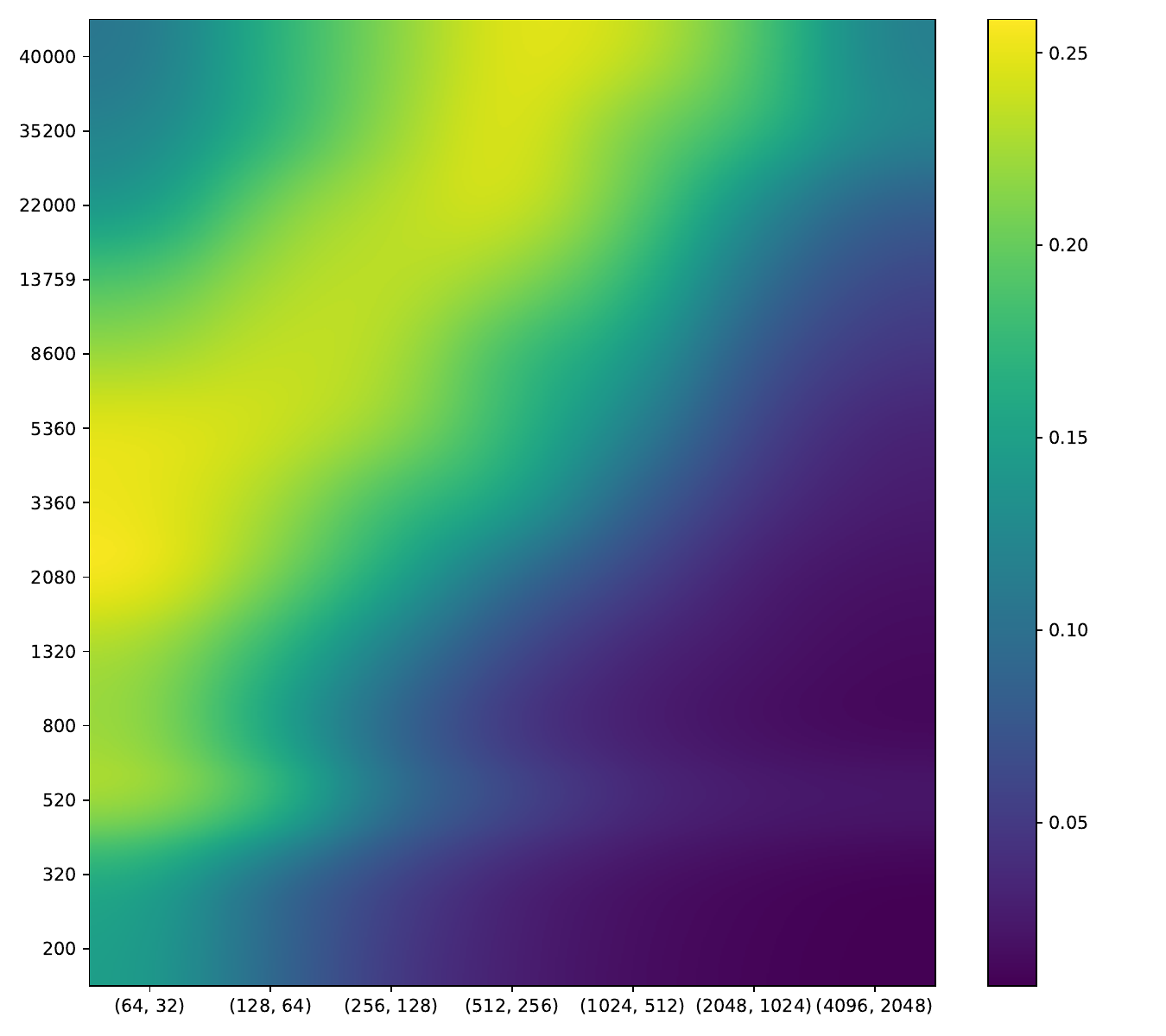}
             \caption{CIFAR10 - OOD}
         \end{subfigure}
         \caption{[Epistemic Uncertainty] Ensemble}
    \end{figure*}

    \newpage
    \section*{AUC ID vs OOD}
    0: ID - 1: OOD - same number of examples for ID and OOD ...
    \begin{figure*}[h!]
         \centering
         \begin{subfigure}[b]{0.16\textwidth}
             \centering
             \includegraphics[width=\textwidth]{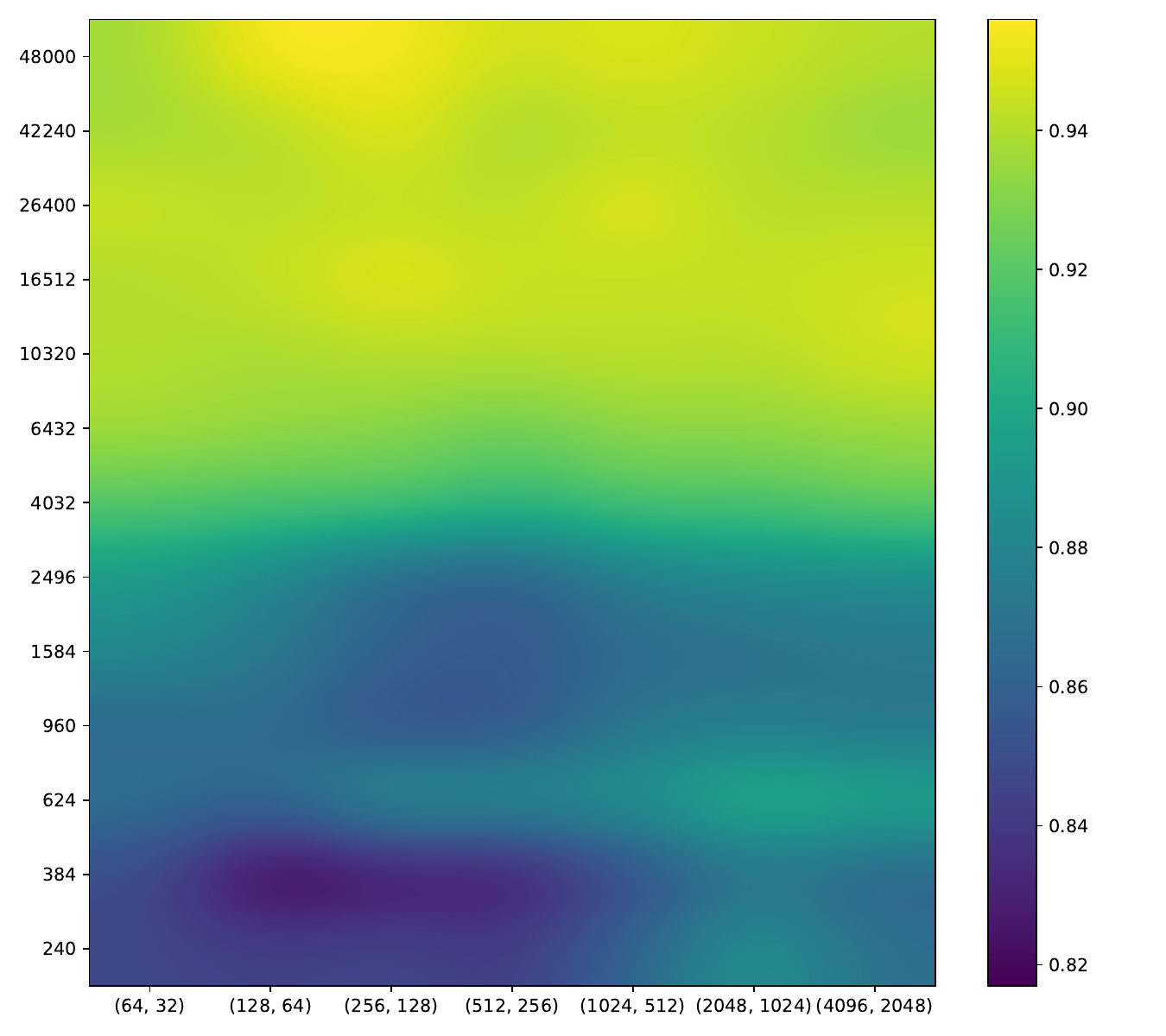}
             \caption{MNIST - Ensemble}
         \end{subfigure}
         \hfill
         \begin{subfigure}[b]{0.16\textwidth}
             \centering
             \includegraphics[width=\textwidth]{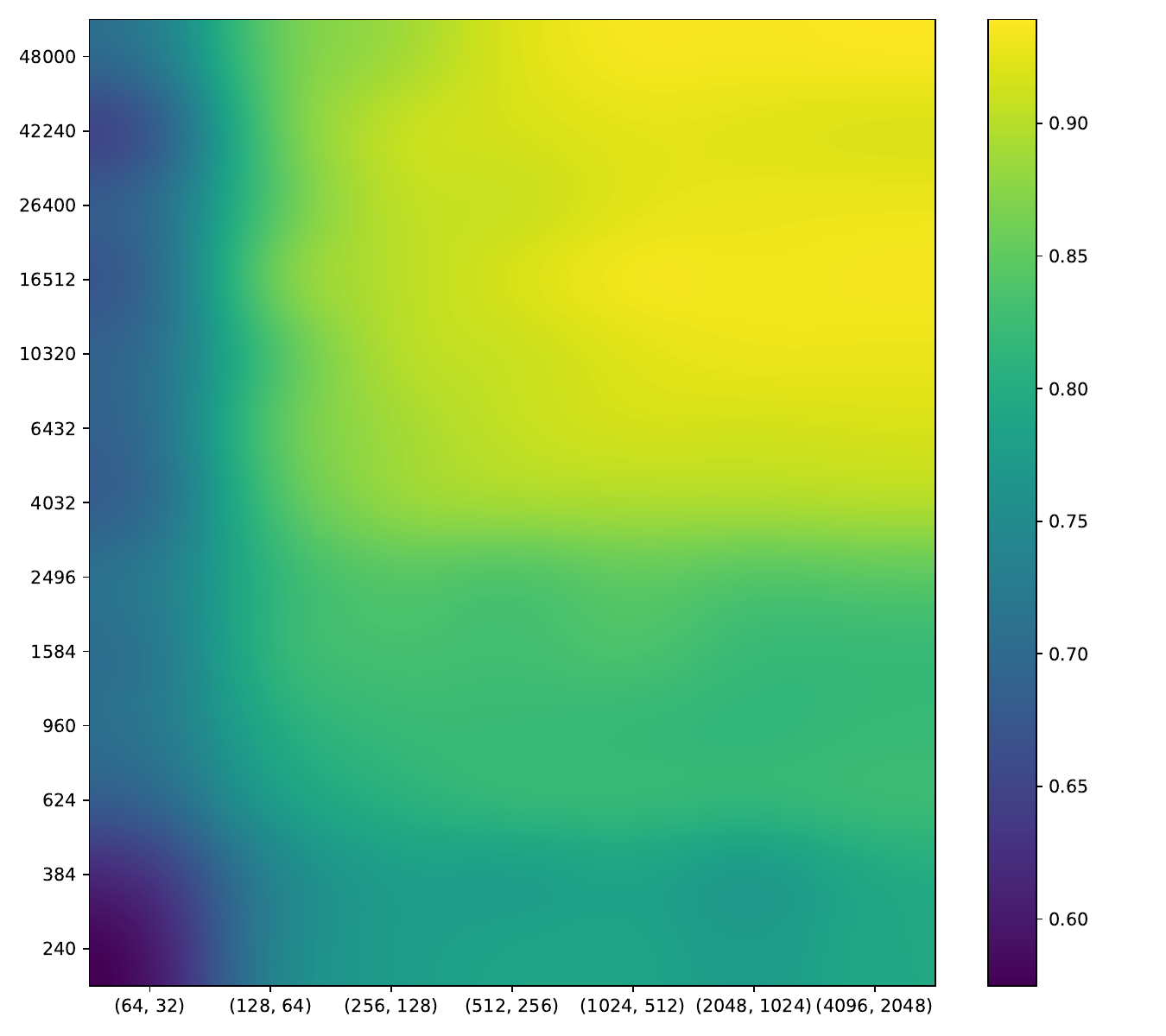}
             \caption{MNIST - MC-Dropout $0.5 - 0.5$}
         \end{subfigure}
         \hfill
         \begin{subfigure}[b]{0.16\textwidth}
             \centering
             \includegraphics[width=\textwidth]{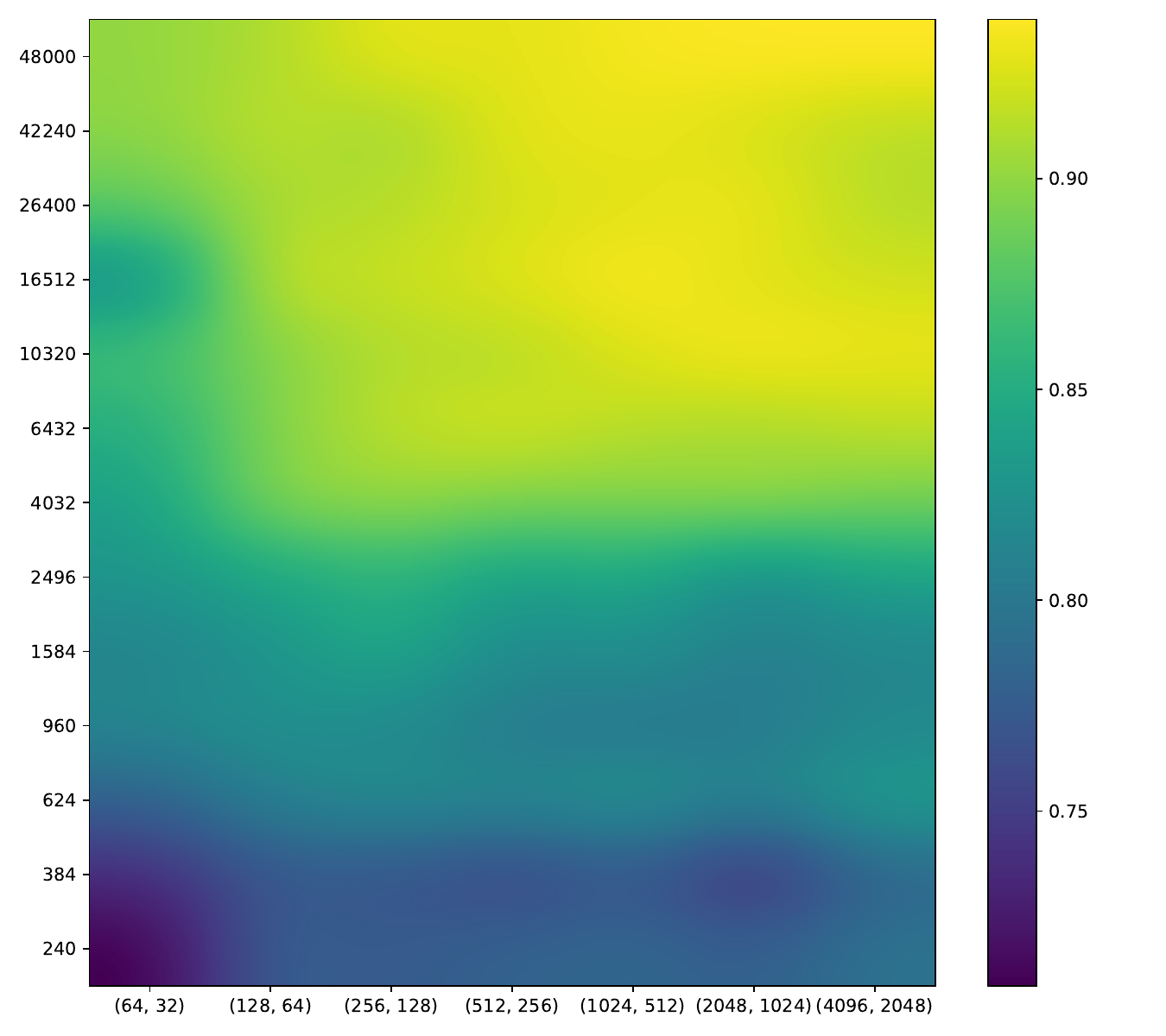}
             \caption{MNIST - MC-Dropout $0.5 - 0.1$}
         \end{subfigure}
         \hfill
         \begin{subfigure}[b]{0.16\textwidth}
             \centering
             \includegraphics[width=\textwidth]{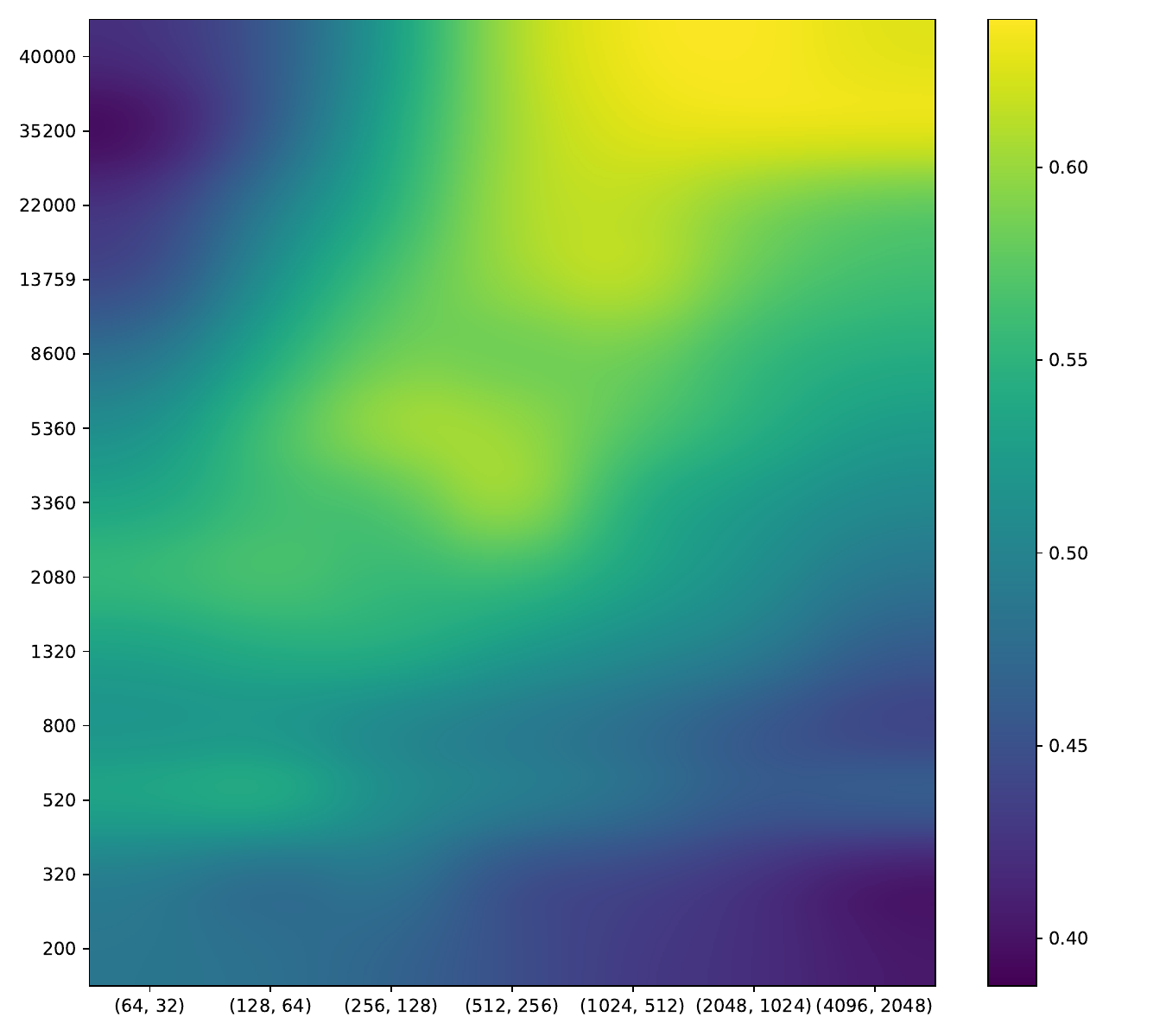}
             \caption{CIFAR10 - Ensemble}
         \end{subfigure}
         \hfill
         \begin{subfigure}[b]{0.16\textwidth}
             \centering
             \includegraphics[width=\textwidth]{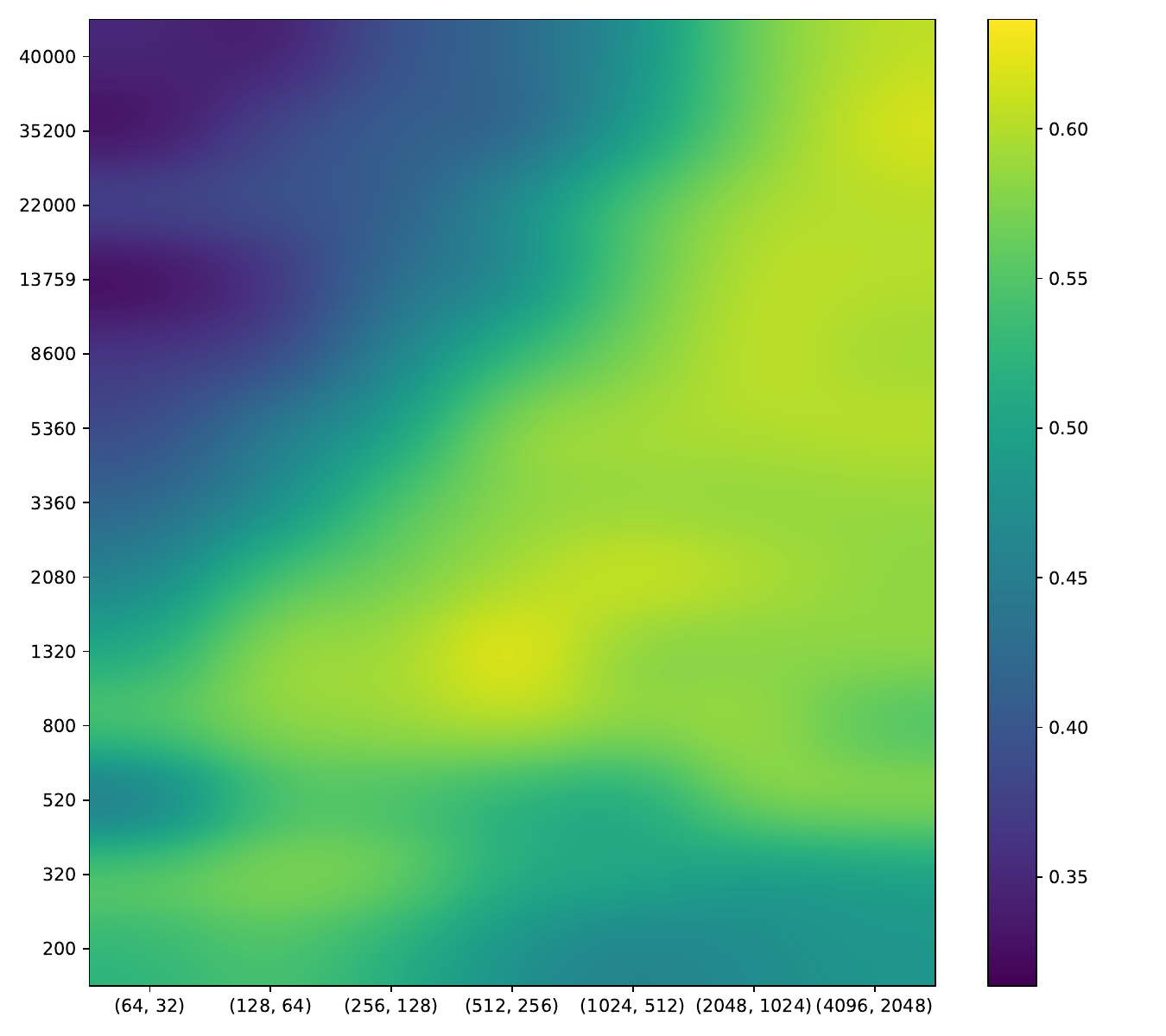}
             \caption{CIFAR10 - MC-Dropout $0.5 - 0.5$}
         \end{subfigure}
         \hfill
         \begin{subfigure}[b]{0.16\textwidth}
             \centering
             \includegraphics[width=\textwidth]{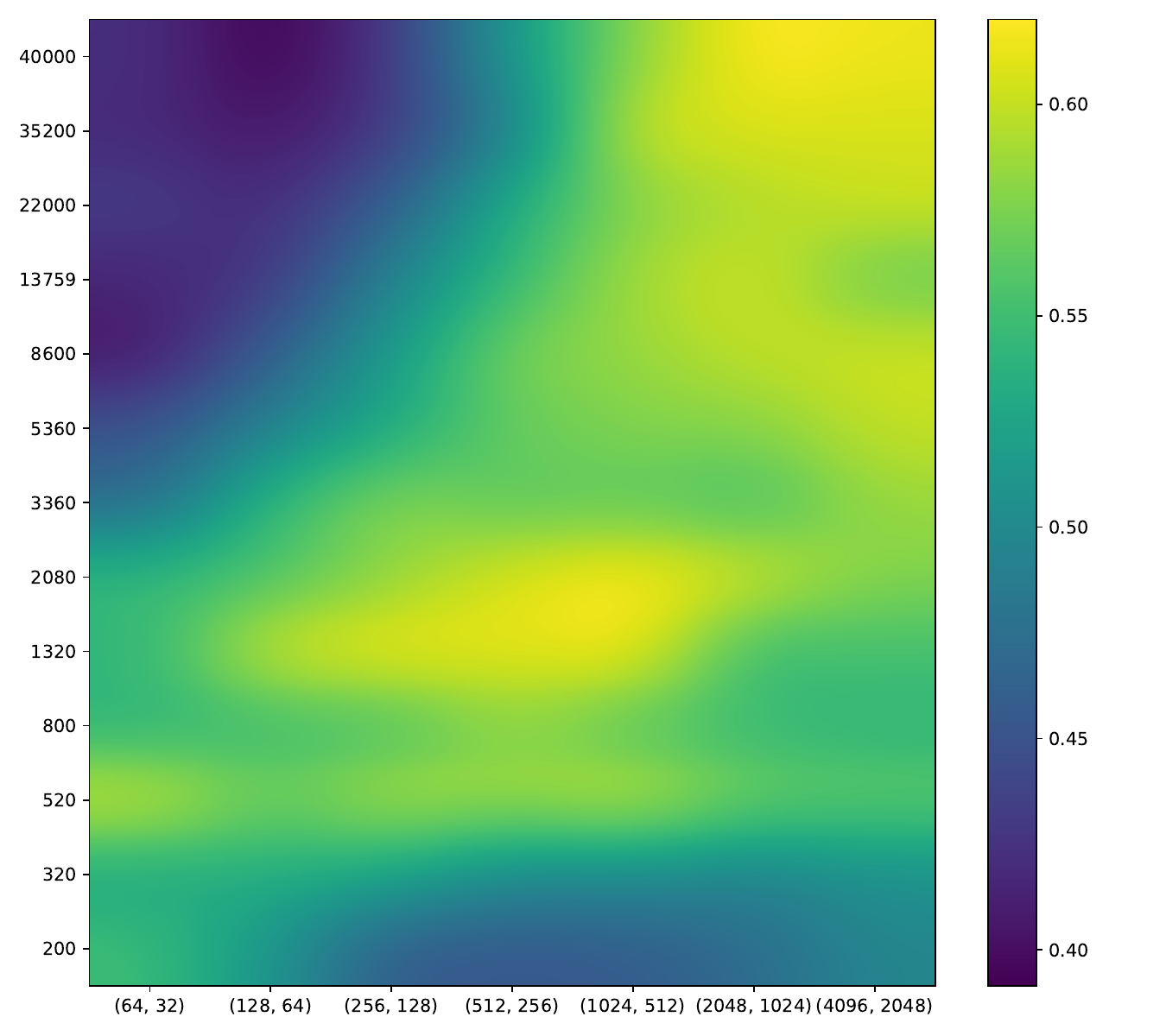}
             \caption{MNIST - MC-Dropout $0.5 - 0.1$}
         \end{subfigure}
         \caption{AUC - Epistemic Uncertainty}
    \end{figure*}
    
    \begin{figure*}[h!]
         \centering
         \begin{subfigure}[b]{0.16\textwidth}
             \centering
             \includegraphics[width=\textwidth]{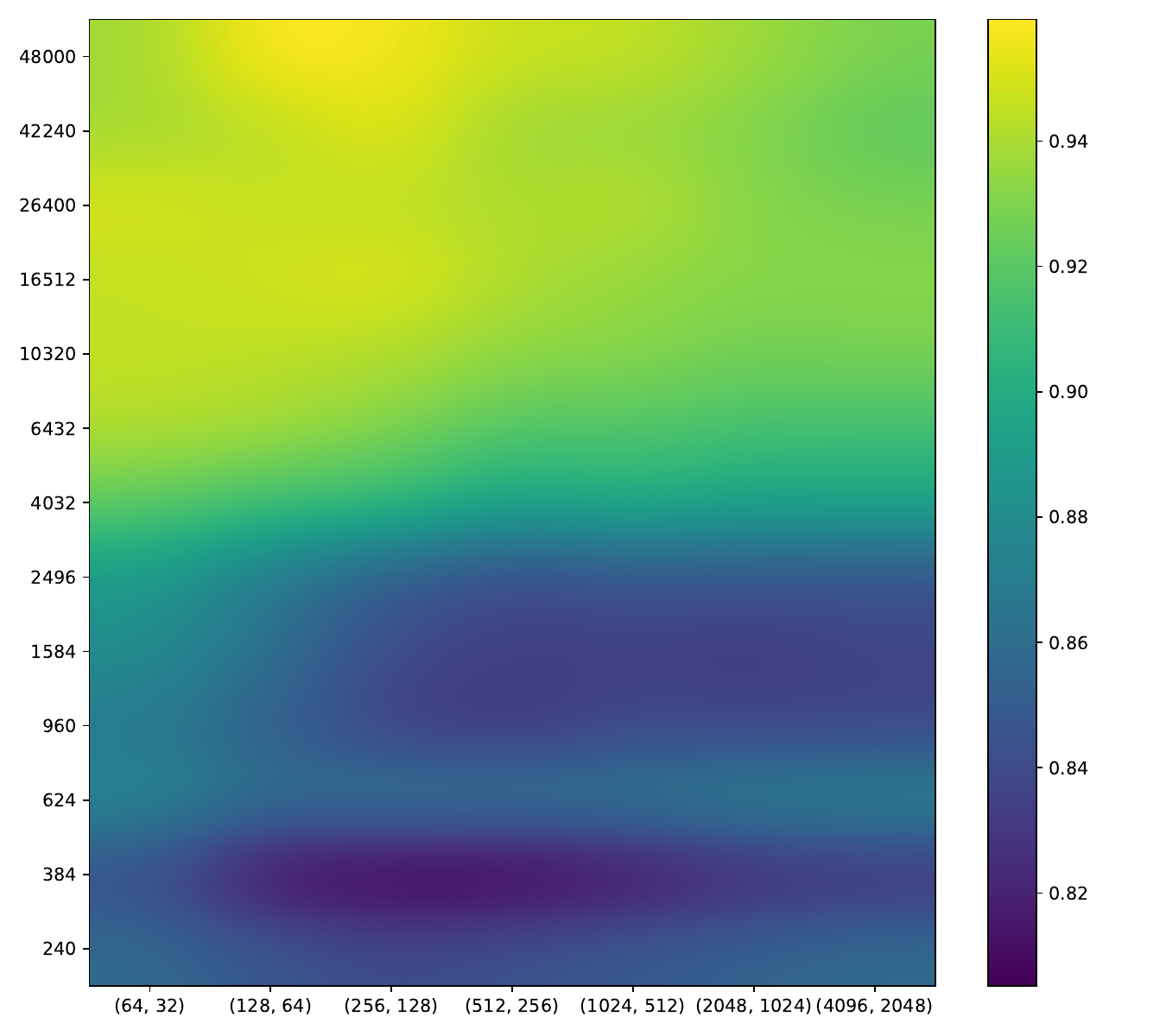}
             \caption{MNIST - Ensemble}
         \end{subfigure}
         \hfill
         \begin{subfigure}[b]{0.16\textwidth}
             \centering
             \includegraphics[width=\textwidth]{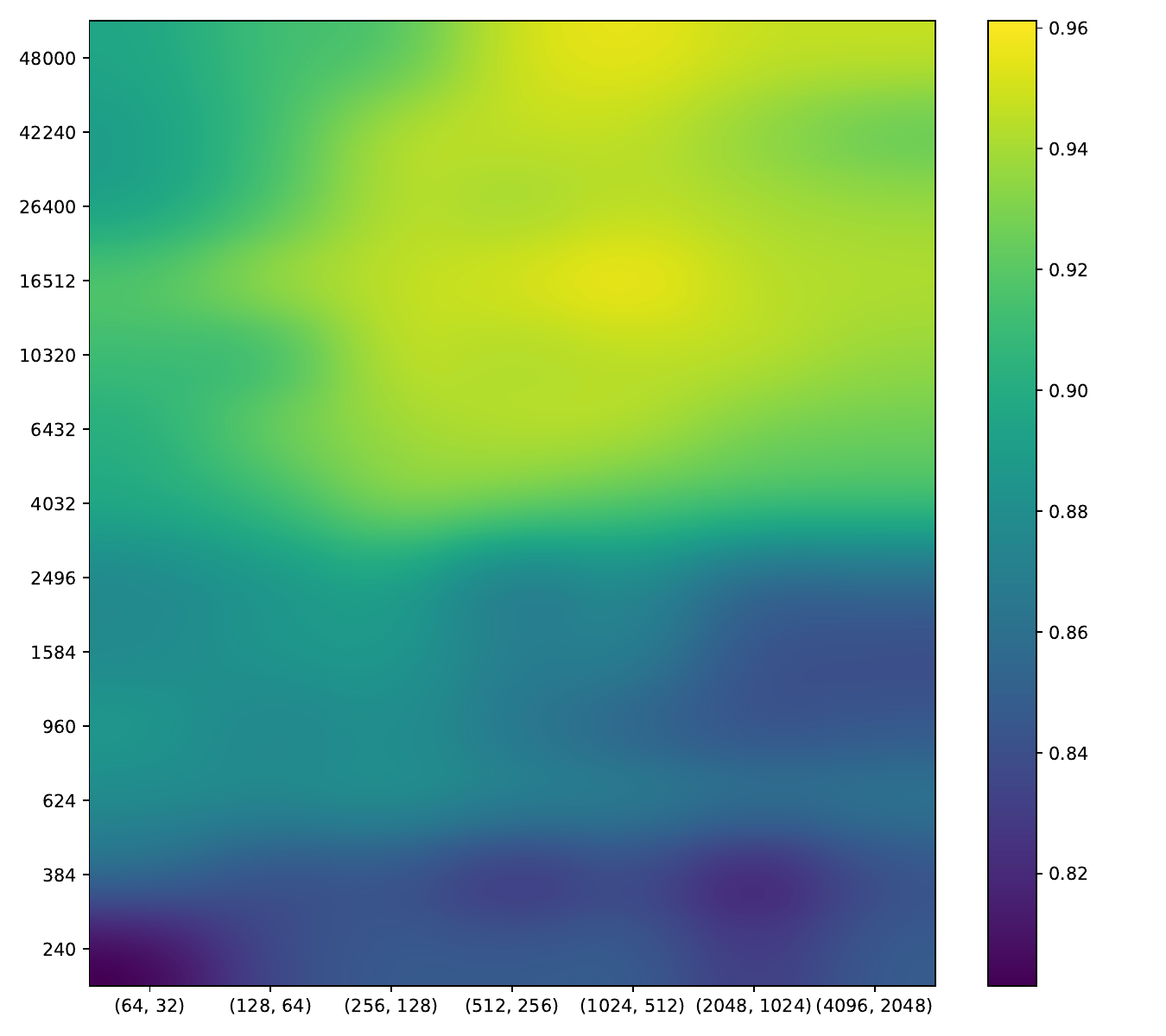}
             \caption{MNIST - MC-Dropout $0.5 - 0.5$}
         \end{subfigure}
         \hfill
         \begin{subfigure}[b]{0.16\textwidth}
             \centering
             \includegraphics[width=\textwidth]{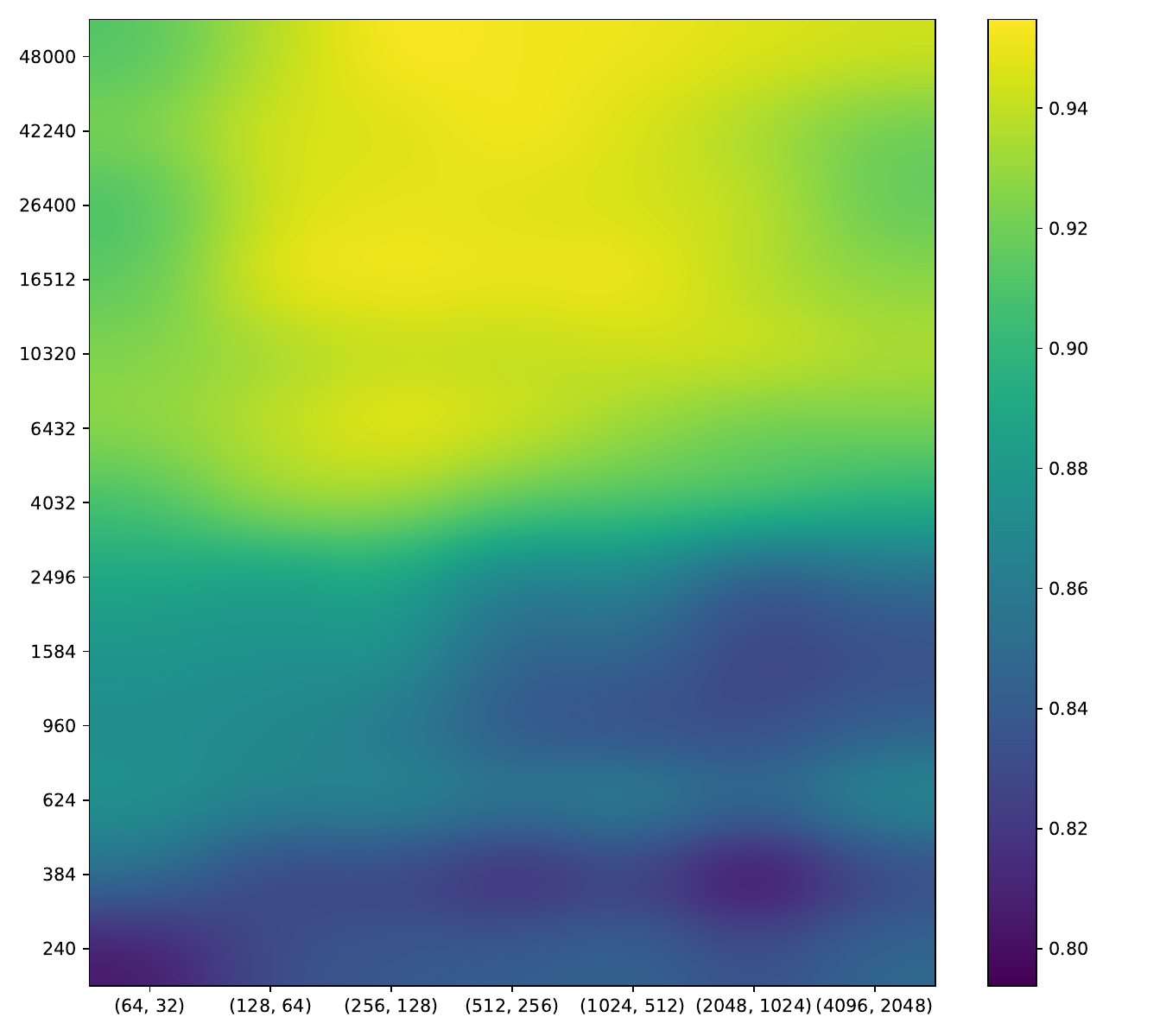}
             \caption{MNIST - MC-Dropout $0.5 - 0.1$}
         \end{subfigure}
         \hfill
         \begin{subfigure}[b]{0.16\textwidth}
             \centering
             \includegraphics[width=\textwidth]{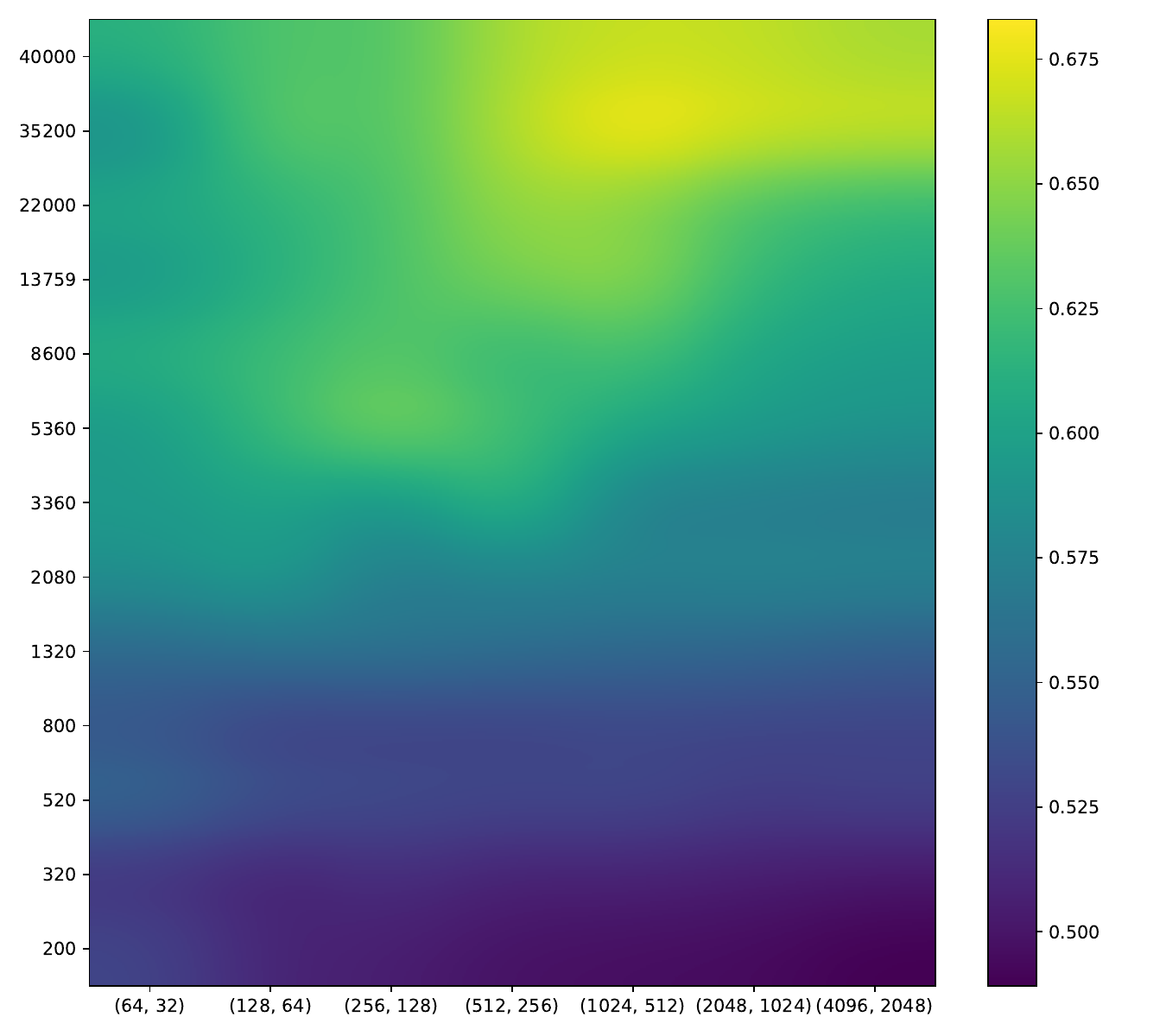}
             \caption{CIFAR10 - Ensemble}
         \end{subfigure}
         \hfill
         \begin{subfigure}[b]{0.16\textwidth}
             \centering
             \includegraphics[width=\textwidth]{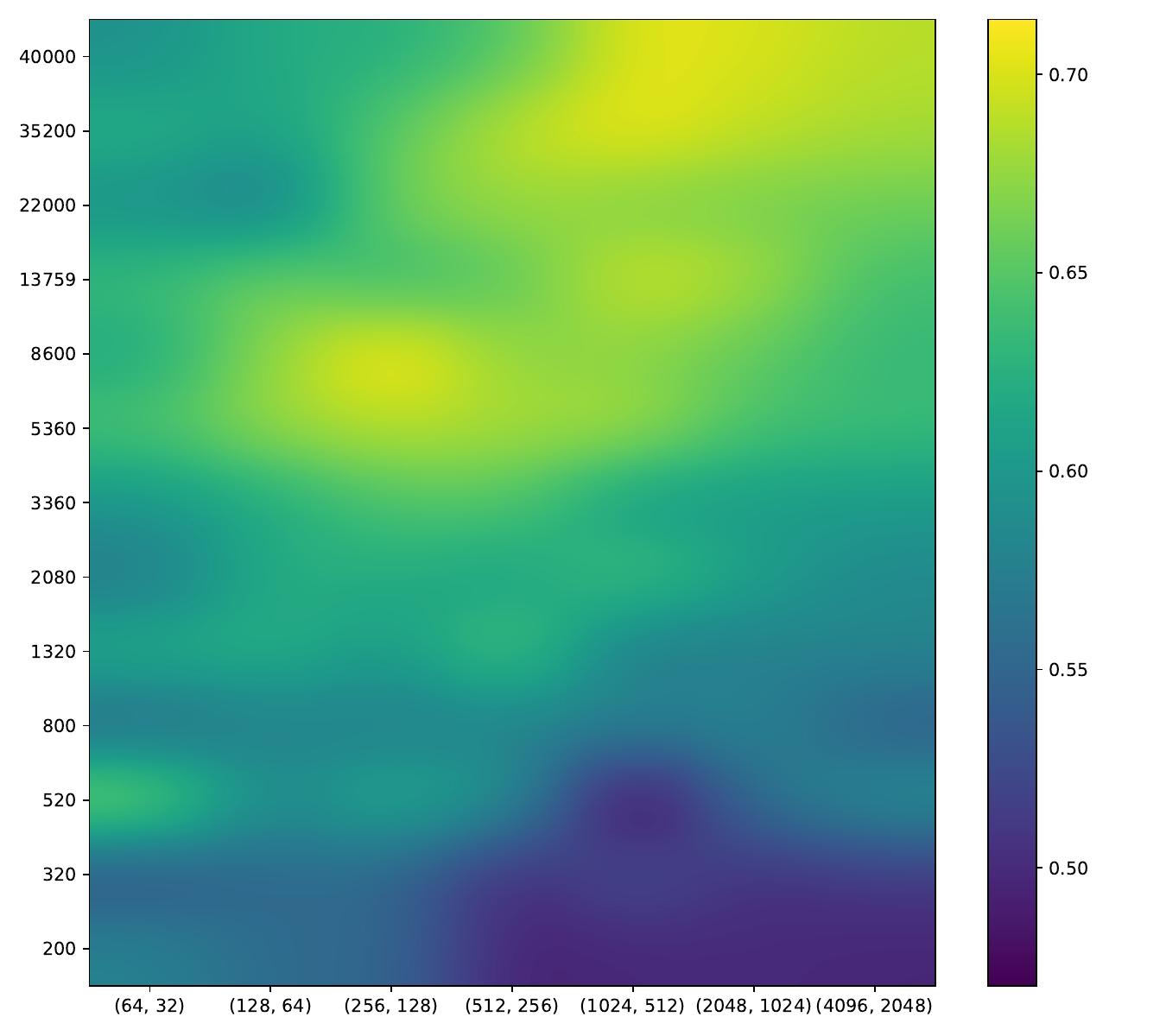}
             \caption{CIFAR10 - MC-Dropout $0.5 - 0.5$}
         \end{subfigure}
         \hfill
         \begin{subfigure}[b]{0.16\textwidth}
             \centering
             \includegraphics[width=\textwidth]{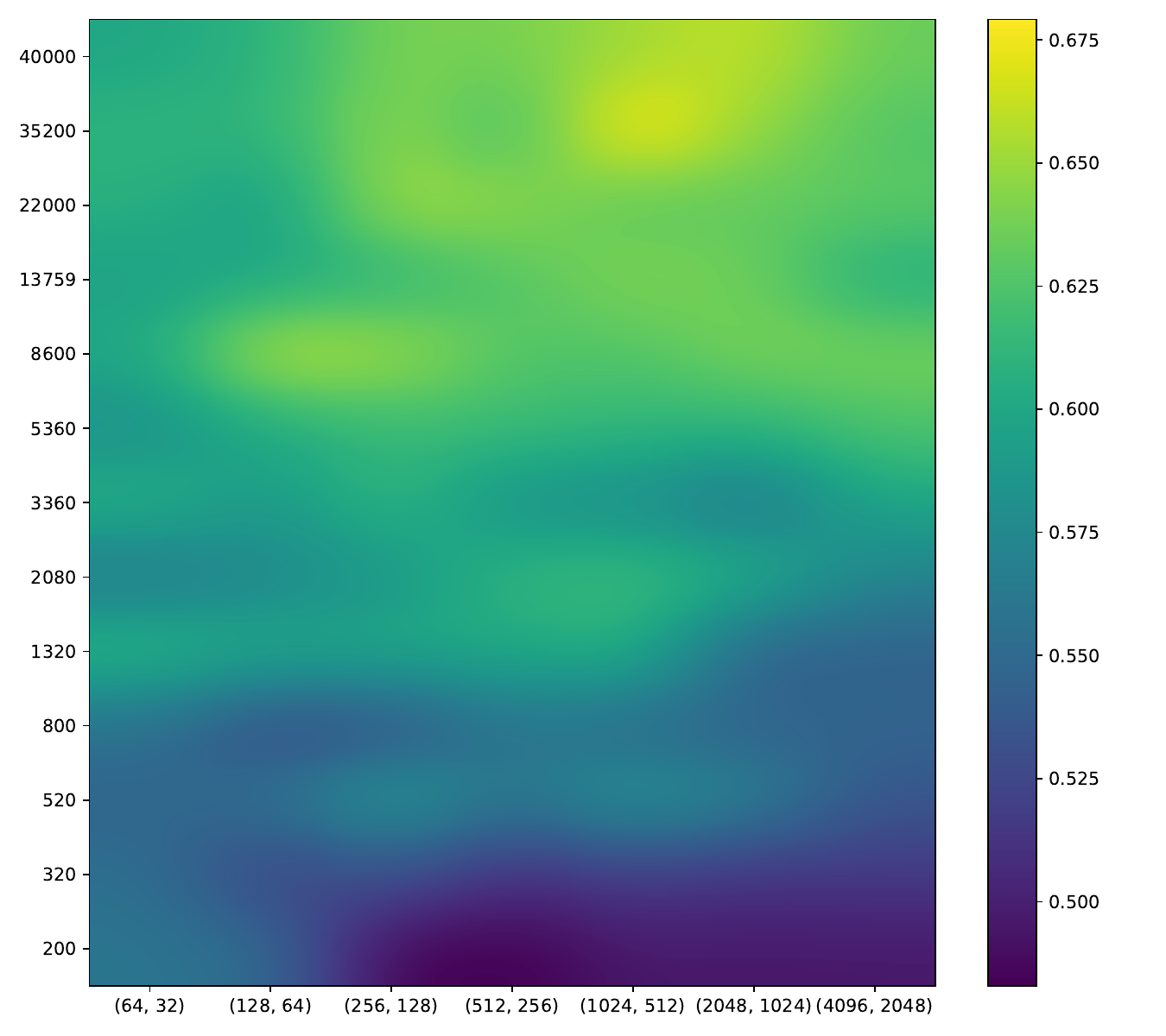}
             \caption{MNIST - MC-Dropout $0.5 - 0.1$}
         \end{subfigure}
         \caption{AUC - Predictive Uncertainty}
    \end{figure*}
    
    \begin{figure*}[h!]
         \centering
         \begin{subfigure}[b]{0.16\textwidth}
             \centering
             \includegraphics[width=\textwidth]{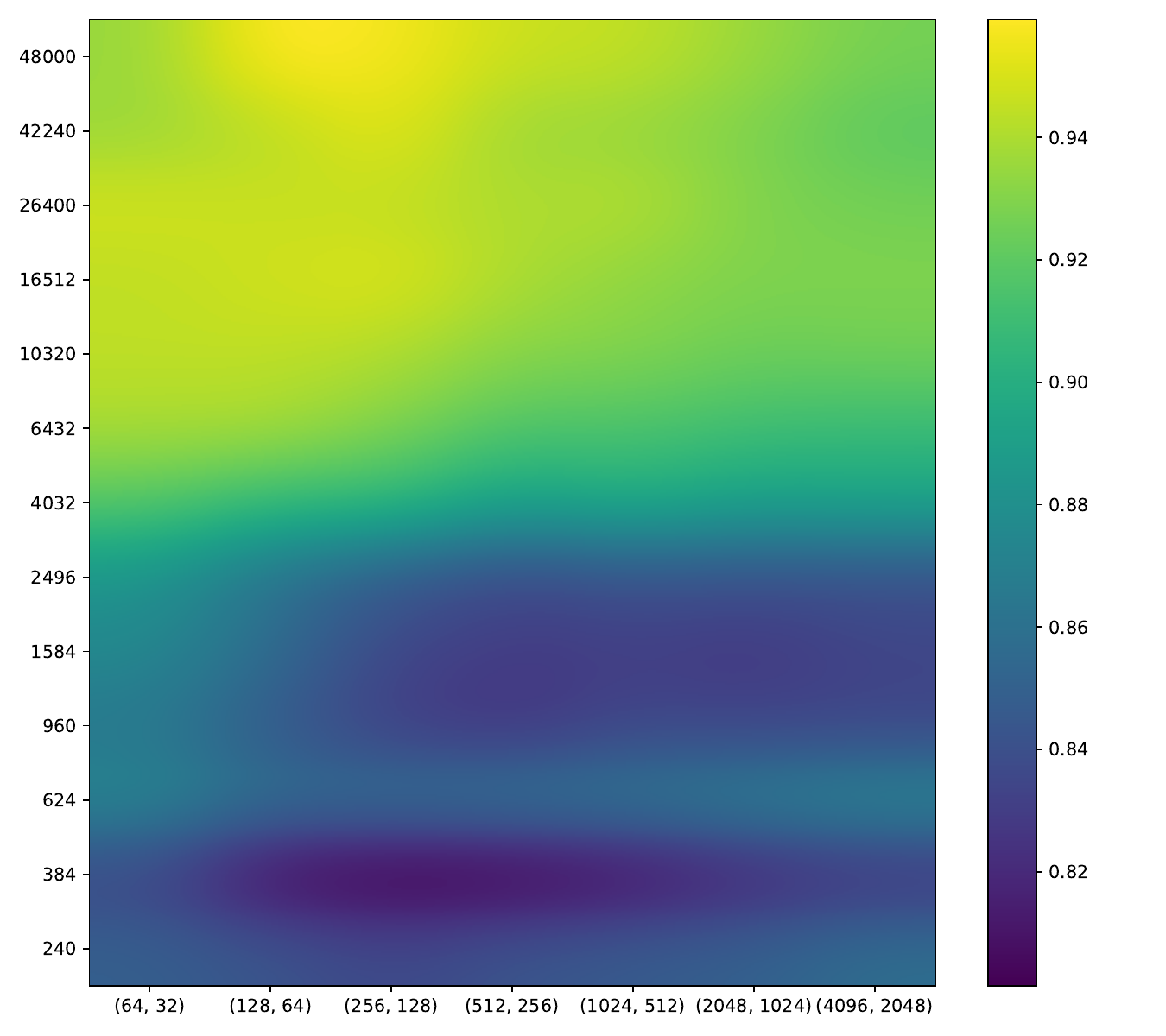}
             \caption{MNIST - Ensemble}
         \end{subfigure}
         \hfill
         \begin{subfigure}[b]{0.16\textwidth}
             \centering
             \includegraphics[width=\textwidth]{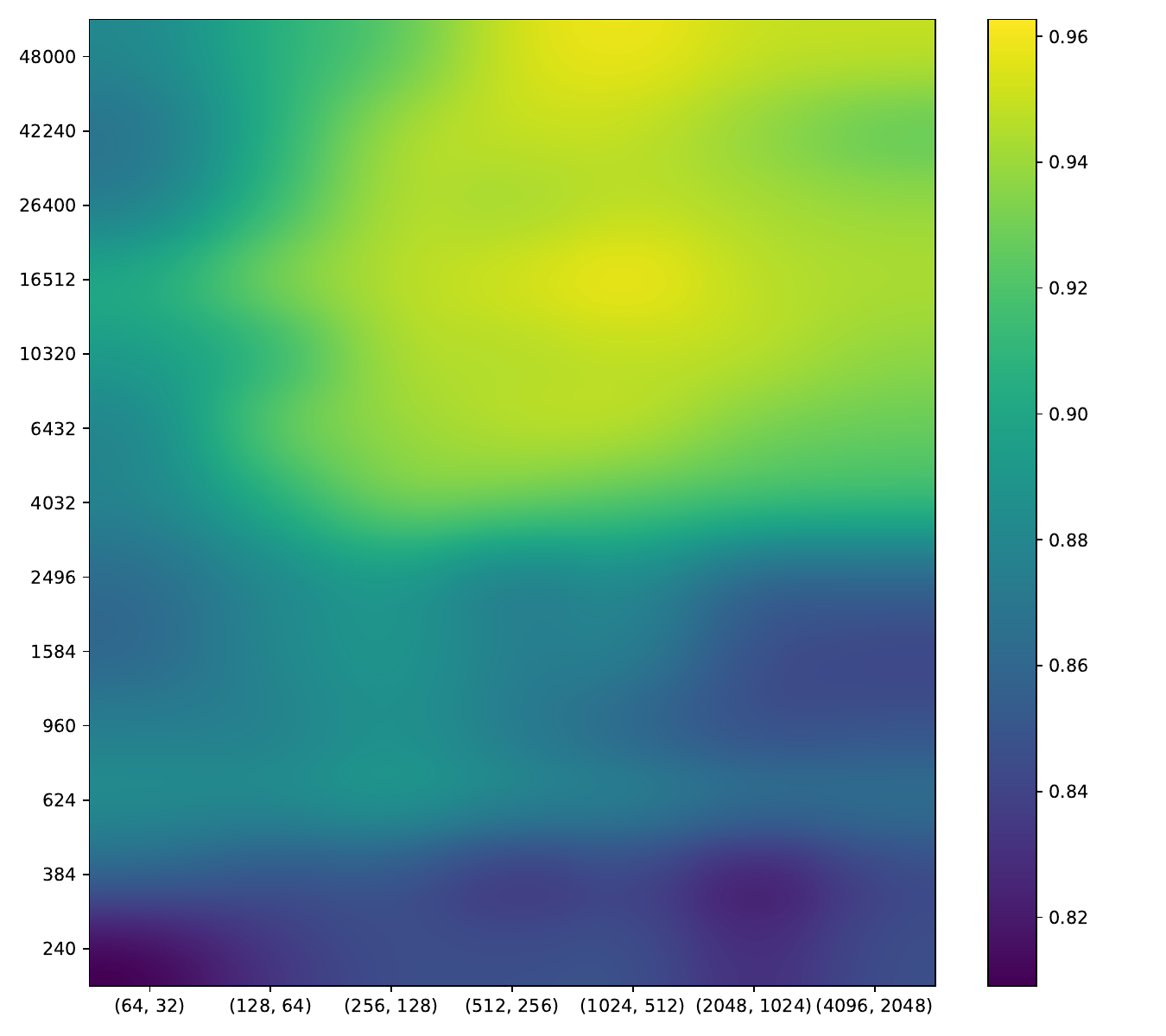}
             \caption{MNIST - MC-Dropout $0.5 - 0.5$}
         \end{subfigure}
         \hfill
         \begin{subfigure}[b]{0.16\textwidth}
             \centering
             \includegraphics[width=\textwidth]{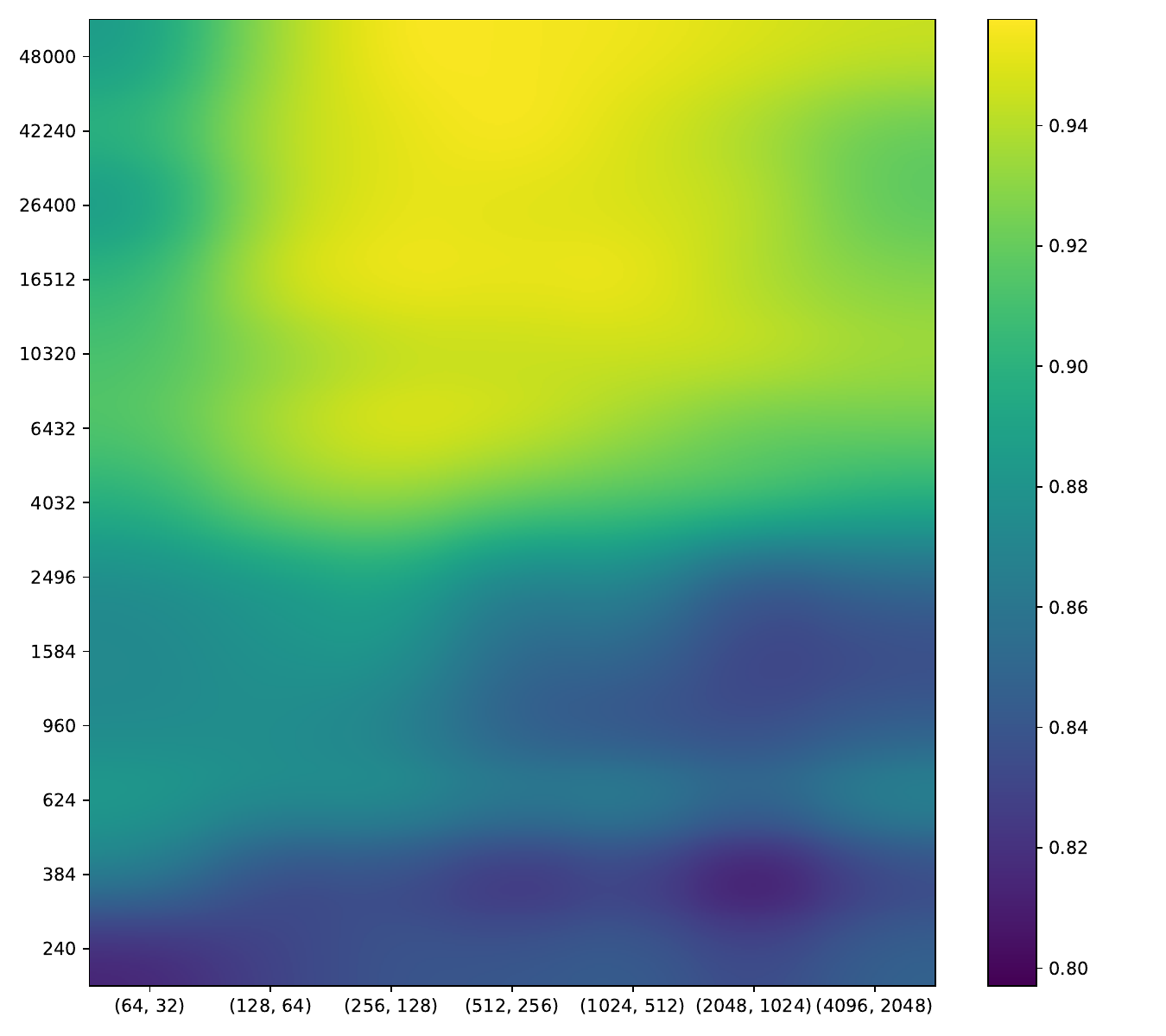}
             \caption{MNIST - MC-Dropout $0.5 - 0.1$}
         \end{subfigure}
         \hfill
         \begin{subfigure}[b]{0.16\textwidth}
             \centering
             \includegraphics[width=\textwidth]{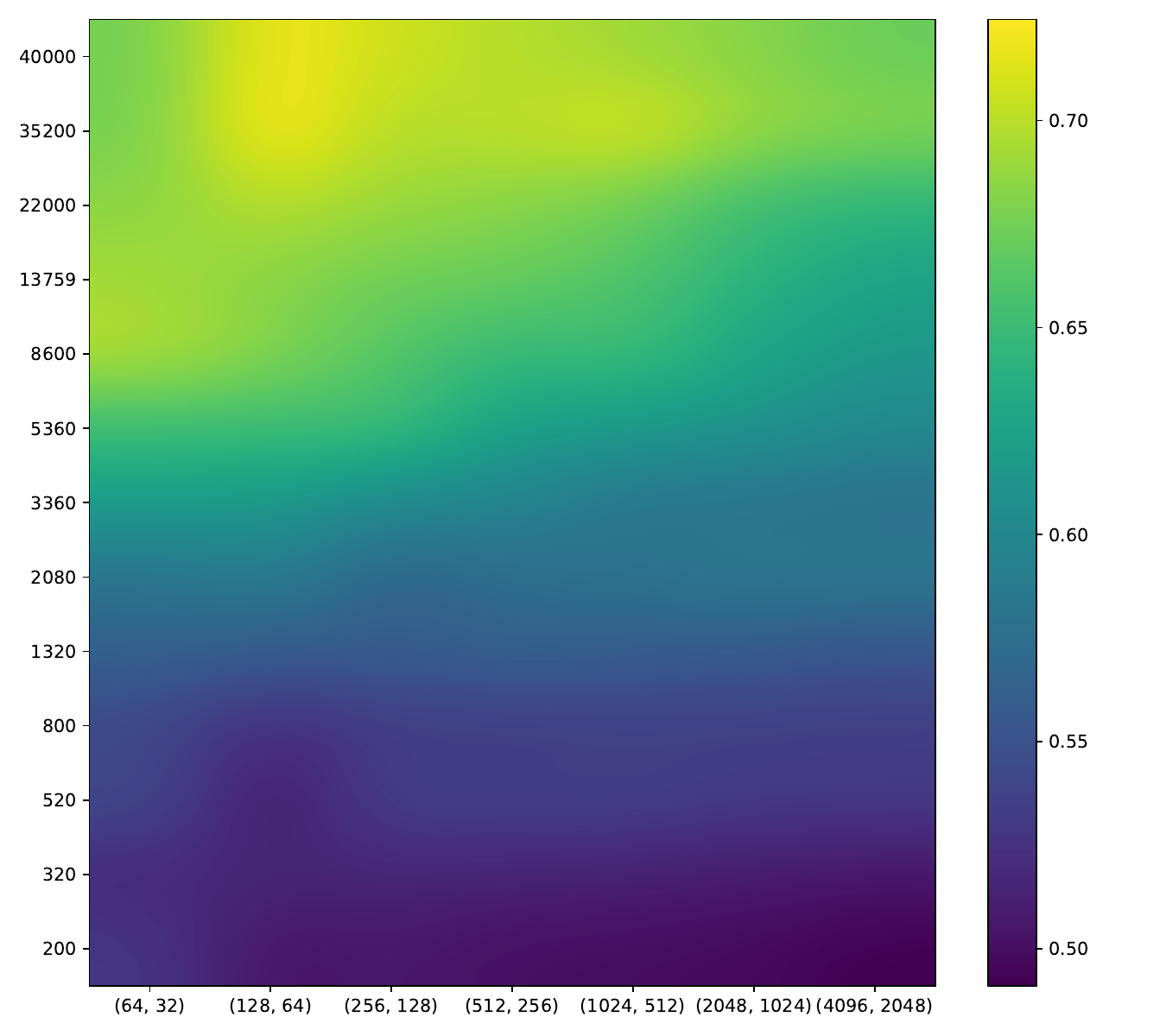}
             \caption{CIFAR10 - Ensemble}
         \end{subfigure}
         \hfill
         \begin{subfigure}[b]{0.16\textwidth}
             \centering
             \includegraphics[width=\textwidth]{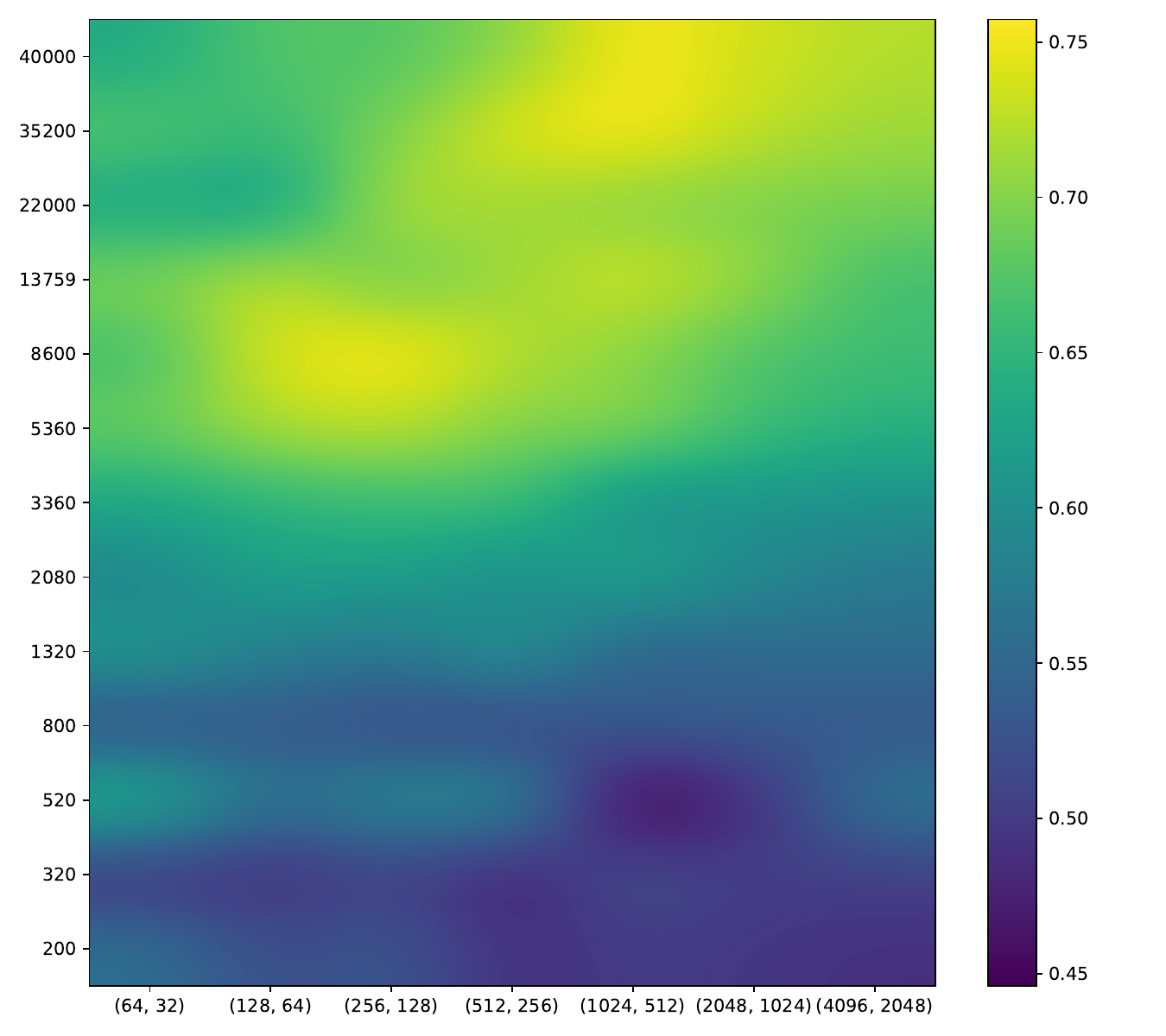}
             \caption{CIFAR10 - MC-Dropout $0.5 - 0.5$}
         \end{subfigure}
         \hfill
         \begin{subfigure}[b]{0.16\textwidth}
             \centering
             \includegraphics[width=\textwidth]{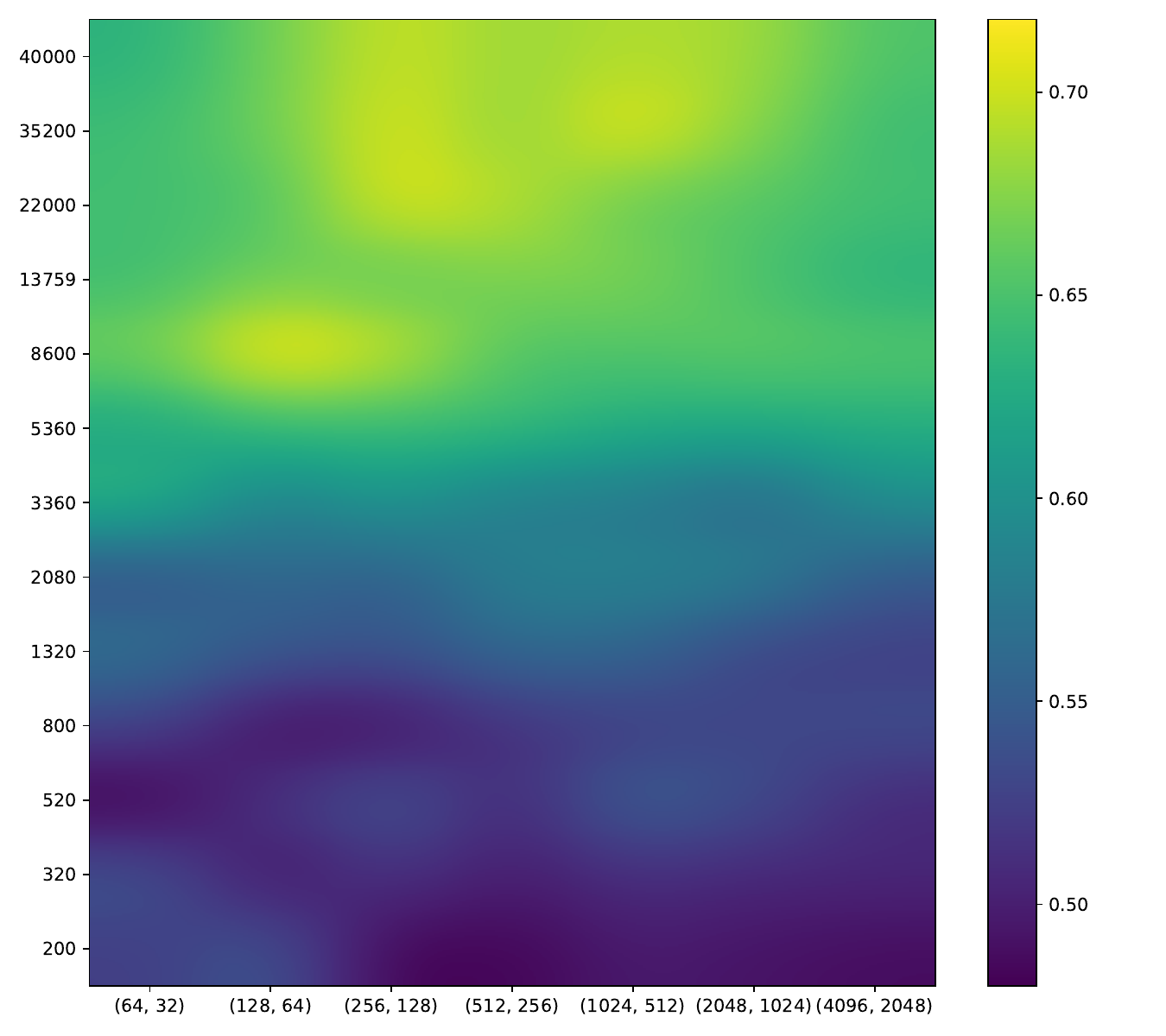}
             \caption{MNIST - MC-Dropout $0.5 - 0.1$}
         \end{subfigure}
         \caption{AUC - Aleatoric Uncertainty}
    \end{figure*}
}

\end{document}